\documentclass{article}

\usepackage{arxiv}

\usepackage[utf8]{inputenc} 
\usepackage[T1]{fontenc}    
\usepackage{hyperref}       
\usepackage{url}            
\usepackage{booktabs}       
\usepackage{amsfonts}       
\usepackage{nicefrac}       
\usepackage{microtype}      
\usepackage{lipsum}		
\usepackage{graphicx}
\graphicspath{{Figures/}}
\usepackage{natbib}
\usepackage{doi}
\usepackage{amsmath,amssymb}
\usepackage{bm}
\usepackage{comment}
\usepackage{xurl}

\usepackage{comment}
\usepackage{xurl}
\usepackage{algpseudocode}
\usepackage[noline,ruled,algo2e,linesnumbered]{algorithm2e}
\usepackage{subfigure}

\usepackage{todonotes}

\DeclareMathOperator*{\argmax}{arg\,max}
\DeclareMathOperator*{\argmin}{arg\,min}

\title{Quantifying uncertainty for deep learning based forecasting and flow-reconstruction using neural architecture search ensembles}

\author{ 
    Romit Maulik\thanks{Email: \texttt{rmaulik@anl.gov}} \\
	Argonne National Laboratory,\\
	Lemont, Illinois 60439, USA.
	\And
	Romain Egele \\
    Universit\'{e} Paris-Saclay, \\
	1190 Gif-sur-Yvette, France.
    \And
	Krishnan Raghavan \\
    Argonne National Laboratory,\\
	Lemont, Illinois 60439, USA.
    \And
	Prasanna Balaprakash \\
    Argonne National Laboratory,\\
	Lemont, Illinois 60439, USA.
}

\renewcommand{\shorttitle}{Neural architecture search ensembles for scientific machine learning}

\begin{document}
\maketitle

\begin{abstract}
Classical problems in computational physics such as data-driven forecasting and signal reconstruction from sparse sensors have recently seen an explosion in deep neural network (DNN) based algorithmic approaches. However, most DNN models do not provide uncertainty estimates, which are crucial for establishing the trustworthiness of these techniques in downstream decision making tasks and scenarios. In recent years, ensemble-based methods have achieved significant success for the uncertainty quantification in DNNs on a number of benchmark problems. However, their performance on real-world applications remains under-explored. In this work, we present an automated approach to DNN discovery and demonstrate how this may also be utilized for ensemble-based uncertainty quantification. Specifically, we propose the use of a scalable neural and hyperparameter architecture search for discovering an ensemble of DNN models for complex dynamical systems. We highlight how the proposed method not only discovers high-performing neural network ensembles for our tasks, but also quantifies uncertainty seamlessly. This is achieved by using genetic algorithms and Bayesian optimization for sampling the search space of neural network architectures and hyperparameters. Subsequently, a model selection approach is used to identify candidate models for an ensemble set construction. Afterwards, a variance decomposition approach is used to estimate the uncertainty of the predictions from the ensemble. We demonstrate the feasibility of this framework for two tasks - forecasting from historical data and flow reconstruction from sparse sensors for the sea-surface temperature.  We demonstrate superior performance from the ensemble in contrast with individual high-performing models and other benchmarks.
\end{abstract}


\section{Introduction}

\subsection{Motivation}

Data-driven surrogate modeling research has shown great promise in improving the predictability and efficiency of computational physics applications. Among various algorithms, deep learning-based models have been observed to show significant gains in accuracy and time-to-solution over classical numerical-methods based techniques. However, the widespread adoption of deep learning models is still limited by its black-box nature. 
To that end, uncertainty quantification methods have been developed to overcome the challenges associated with the black-box nature of the deep-learning models and establish trustworthiness by providing uncertainty estimates along with the predictions. The data or aleatoric uncertainty is attributed to the noise in the data, for example, low resolution sensors and sparse measurements; the model or epistemic uncertainty is attributed to the lack of training data. The former is inherent to the data and cannot be reduced by collecting more data; the latter is used to characterize the model's predictive capability with respect to the training data and it can reduced by collecting appropriate data. In many computational physics applications, it is crucial to effectively quantify both data- and model-form uncertainties in predictions from deep learning models. Quantification of such uncertainties alleviates the risks associated with the deployment of such black-box data-driven models for real-world tasks.

Deep ensembles is a promising approach for uncertainty quantification. In this approach, a ensemble of neural networks (NNs) are trained independently but they differ in the way in which they are trained. Consequently, the weights of the neural network parameters will be different. The prediction from these models are then used to improve prediction and estimate uncertainty. Despite its simplicity, deep-ensembles-based uncertainty estimation has achieved superior performance over more sophisticated uncertainty quantification methods on a number of benchmarks. 

Crucial to the effectiveness of uncertainty quantification in the deep-ensemble-based methods is the diversity of the high-performing models. Specifically, if the models are significantly different from each other and also equally high performing, then the prediction and the uncertainty estimates become accurate \citep{egele2022autodeuq}. However, this poses additional challenges due to the large overhead associated with the manual design of models. Despite one time cost, the development overhead significantly increases the offline design and training costs of the ensemble model development and increase amortization time (i.e., the time required to offset offline costs).
To that end, we explore an integrated automated deep ensemble approach that not only discovers high-performing models but also makes ensemble predictions with quantified uncertainty in a scalable manner. Specifically, the highlights of this article are as follows:
\begin{itemize}
    \item We demonstrate a unified strategy for discovering high-performing deep learning models for dynamical systems forecasting and signal recovery with epistemic and aleatoric uncertainty quantification that leverages distributed computing. 
    \item For uncertainty quantification, we use the law of decomposition of variance from members of an ensemble to separately estimate both the aleatoric and epistemic uncertainty.
    \item We validate our proposed approach for forecasting flow-fields as well as instantaneous state reconstructions for a real-world, high-dimensional scientific machine learning problem with complex dynamics, given by the NOAA Optimum Interpolation Data Set. 
\end{itemize}

\subsection{Related work}

In recent times, deep learning models have been popular for complex predictive modeling tasks. In this section, we review past work in surrogate modeling of dynamical systems as well as attempts to quantify the uncertainty of these data-driven models. Data-driven surrogate models are primarily designed to reduce the computational costs associated with expensive forward models in many-query tasks \citep{carlberg2011efficient,wang2012proper,san2015principal,ballarin2015supremizer,san2018extreme,wang2019non,choi2019space,renganathan2020machine,ren2020lower}. Therefore these find extensive application  in control \citep{proctor2016dynamic,peitz2019multiobjective,noack2011reduced,rowley2017model,raibaudo2020machine,ren2020active}, optimization \citep{peherstorfer2016optimal,fan2020optimization}, uncertainty quantification \citep{sapsis2013statistically,zahr2018efficient,goh2019solving} and data-assimilation \citep{arcucci2019optimal,tang2020deep,casas2020reduced,maulik2022efficient} among others. Typically, for complex dynamical systems possessing large degrees of freedom, a first step in model construction is the identification of a reduced-basis for evolving dynamics \citep{san2014basis,korda2018data,kalb2007intrinsic}. After identifying this basis, a cost-effective strategy is necessary for evolving the dynamics in this transformed space \citep{kalashnikova2010stability,mohebujjaman2019physically,carlberg2011efficient,xiao2013non,fang2013non,carlberg2017galerkin,huang2022accelerating}. A popular approach to this latent-space dynamics evolution has been the use of neural networks with inductive biases for sequential data. For example, long short-term memory (LSTM) neural networks \citep{hochreiter1997long} have recently become very popular in modeling these dynamics  \citep{vlachas2018data,ahmed2019long,maulik2019time,mohan2019compressed,mohan2018deep,gonzalez2018learning,hasegawa2020cnn,wang2020recurrent,hasegawa2020machine,chattopadhyay2020data,maulik2020non}. Other methods include Gaussian processes \citep{renganathan2021enhanced,ma2021data,guo2018reduced}, transformer models \citep{vaswani2017attention,geneva2022transformers} which use temporal attention to learn patterns and make forecasts as well as neural ordinary differential equations (commonly known as the neural ODE), where neural networks are assumed to parameterize the right-hand side of an ODE \citep{linot2023stabilized,maulik2019time}. We note that this review is not exhaustive and that several research studies have utilized variants of LSTMs with greater complexity for complex forecasting tasks. 

Data-driven methods have also seen extensive use in flow-reconstruction from sparse sensors. This is due to a desire to build better technologies to reconstruct fluid-flow fields, that may have a large number of frequencies and degrees of freedom from sparse fixed sensors or moving tracer particles (such as those used in experimental imaging with particle image velocimetry). Typically, these techniques rely on linear stochastic \citep{adrian1988stochastic} or gappy proper orthogonal decomposition \citep{everson1995karhunen} estimation. The former reconstructs the flow-field by building a correlation matrix between the sensor inputs and the full flow-fields while the latter solves a linear least-squares problem in an affine subspace spanned by the truncated proper orthogonal decomposition basis vectors. Neural network architectures have been seen to outperform these classical methods for various applications \citep{fukami2020assessment,carter2021data,erichson2020shallow}. Variational inference-based neural architectures, potentially augmented with physics-informed biases, have also been used to reconstruct flow fields from sparse sensors with quantified uncertainty \citep{sun2020physics,dubois2022machine}. However, it is well-known that Bayesian approaches add significant computational cost, mainly due to their sampling process (e.g. Markov Chain Monte Carlo for characterizing the posterior) and are outperformed by ensemble-based approaches for uncertainty quantification \citep{egele2022autodeuq}.

From the perspective of uncertainty quantification for machine learning applications of computational science, several recent articles have attempted to develop algorithms that provide both predictions as well as confidence interval estimates. Here we draw a distinction between using machine learning to accelerate parameteric uncertainty quantification, for example in Bayesian inversion with surrogate models (examples include \cite{sheriffdeen2019accelerating,constantine2016accelerating,goh2019solving,liu2021uncertainty,lye2020deep}), and performing uncertainty quantification of the machine learning algorithm predictions themselves. In the context of the latter, recently \cite{maulik2020probabilistic} demonstrate how a probabilistic neural network may be used to capture the aleatoric uncertainty associated with surrogate model or flow-reconstruction predictions. In \cite{morimoto2022assessments}, the epistemic uncertainty of deep learning frameworks was captured using a weight sampling procedure during convergence, however, with strong assumptions of approximate convexity of the loss surface. In practical problems, particularly with physics-based constraints which promote non-smoothness \citep{krishnapriyan2021characterizing}, such an approach is limited. In \cite{zhu2019physics}, a probabilistic surrogate model, based on a generative neural network architecture (the normalizing flow) was able to obtain a probability density function of the output quantity of interest, conditioned by the input parameters. While this approach avoids the Gaussian assumption for aleatoric uncertainty in \cite{maulik2020probabilistic}, the weights of the surrogate were deterministic and therefore did not quantify NN epistemic uncertainty. In \cite{pawar2022multi}, both aleatoric and epistemic uncertainty were accounted for using the deep ensembles approach but all members of the ensemble were identical neural network architectures initialized differently and therefore were not optimized for the function approximation task.

In this study, we jointly address the issue of optimal architecture and hyperparameter selection, as well as deep ensembles based uncertainty quantification. It is well-known that selecting the best neural network architecture for a function approximation is a difficult task - to that end we have recently demonstrated how scalable neural architecture search may be used for surrogate model discovery \cite{maulik2020recurrent}. In this article, we further expand on our algorithm by demonstrating how the architecture search may be vertically integrated for ensemble forecasting and signal recovery while providing predictions with confidence intervals.

\section{Methods and data}

\subsection{AutoDEUQ: Uncertainty quantification through deep neural network ensembles}

\begin{figure}
    \centering
    \includegraphics[width=\textwidth]{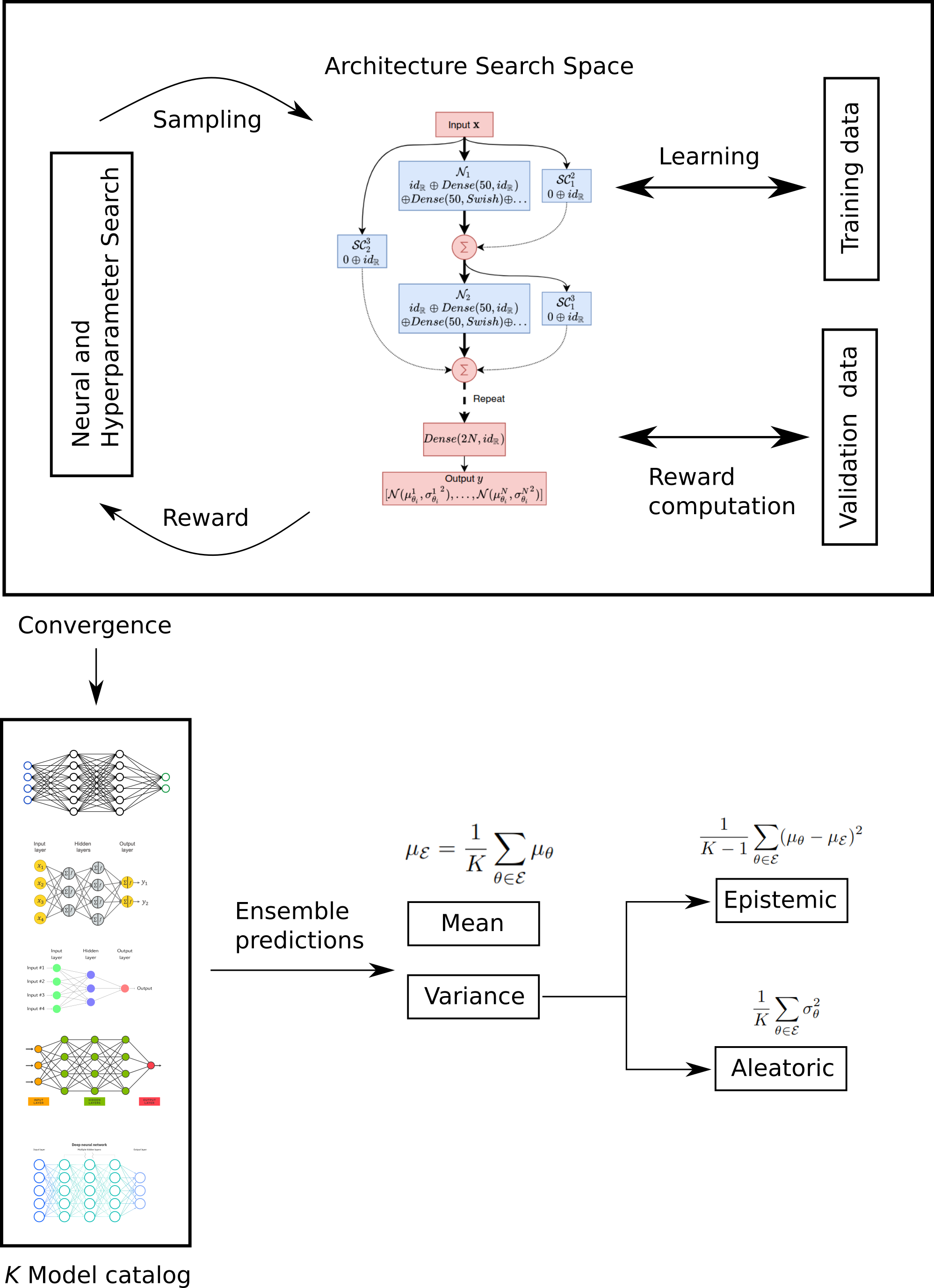}
    \caption{Overall schematic of deep ensembles based uncertainty quantification using AutoDEUQ \cite{egele2022autodeuq}.}
    \label{schematic_de}
\end{figure}

Automated deep ensemble with uncertainty quantification (AutoDEUQ) is a recently proposed  deep-ensemble-based UQ method. AutoDEUQ estimates aleatoric and epistemic uncertainties by automatically generating a catalog of NN models through joint neural architecture and hyperparameter search, wherein each model is trained to minimize the negative log-likelihood to capture aleatoric uncertainty (data noise/uncertainty), and selecting a set of similarly performing models from the catalog to construct the ensembles and model epistemic uncertainty (i.e., function approximation uncertainty). The overall schematic of our neural architecture search based deep ensembles algorithm for uncertainty quantification of scientific machine learning applications is shown in Figure \ref{schematic_de}.

We assume a supervised learning scenario, with a the dataset $\mathcal{D}$ composed of i.i.d points $\left(\mathbf{x}_{i} \in \mathcal{X}, y_{i} \in \mathcal{Y}\right)$, where $\mathbf{x}_{i}$ and $y_{i}=f\left(\mathbf{x}_{i}\right)$ are the input and the corresponding output of the $i$th point, respectively, and $\mathcal{X} \subset \mathbb{R}^N$ and $\mathcal{Y} \subset \mathbb{R}^M$ are the input and output spaces of $N$ and $M$ dimensions, respectively. In the context of regression problems, a focus of our study, the outputs are given by a scalar or vector of real values. Given $\mathcal{D}$, we seek to characterize the predictive distribution $p(y|\mathbf{x})$ using a parameterized distribution $p_\theta(y|\mathbf{x})$, which estimates aleatoric uncertainty through individual trained NNs and then estimates the epistemic uncertainty with an ensemble of high-performing NNs $p_\mathcal{E}(y|\mathbf{x})$. We define $\Theta$ to be the sample space for $\theta.$

The aleatoric uncertainty can be characterized by using the quantiles of $p_{\theta}.$  Following previous work~\cite{lakshminarayanan2016simple}, we make the usual assumption of a Gaussian distribution for $p_{\theta} \sim \mathcal{N}(\mu_\theta, \sigma_\theta^2)$ and use variance as a measure of the aleatoric uncertainty. We explicitly partition $\theta$ into $(\theta_a, \theta_h, \theta_w)$ such that $\Theta$ is decomposed into $(\Theta_a, \Theta_h, \Theta_w)$, where $\theta_a \in \Theta_a$ represents the values of the neural architecture decision variables (network topology parameters), $\theta_h \in \Theta_h$ represents NN training hyperparameters (e.g., learning rate, batch size), and $\theta_w \in \Theta_w$ represents the NN weights. The NN is trained to output mean $\mu_{\theta}$ and variance $\sigma_{\theta}^2$. For a given choice of architecture decision variables $\theta_a$ and training hyperparameters $\theta_h$, to obtain $\theta_w^*,$ we seek to minimize the negative log-likelihood given the real data $\mathcal{D}$. Specifically, we can learn the aleatoric uncertainty using the negative log-likelihood loss (as opposed to the usual mean squared error) in the training: ~\cite{lakshminarayanan2016simple}:
\begin{align}
    \label{eq:ll_func}
    \ell(\mathbf{x}, y; \theta) = -\log p_\theta =\frac{\log \sigma_\theta^{2}(\mathbf{x})}{2}+\frac{\left(y-\mu_{\theta}(\mathbf{x})\right)^{2}}{2 \sigma_{\theta}^{2}(\mathbf{x})}+ \text{cst},
\end{align}
where cst is a constant. The NN training problem is then 
\begin{equation} \label{eq:opt}
    \theta_w^* = \argmax_{\theta_w \in \Theta_w} \quad \ell(\mathbf{x}, y; \theta_a, \theta_h, \theta_w).
\end{equation}

To model epistemic uncertainty, we use deep ensembles~(an ensemble composed of NNs) \cite{lakshminarayanan2016simple}. In our approach, we generate a catalog of NN models $\mathcal{C} = \{\theta_i, i = 1, 2, \cdots, c\}$ (where $\theta \in \Theta$ is a tuple of architecture, optimization hyperparameters, and weights) and repeatedly sample $K$ models to form the ensemble $\mathcal{E} | \mathcal{E} =\{ \theta_i, i=1,2,\cdots,K \}$. 
Let $p_{\theta}$ describe the probability that $\theta$ is a member of the ensemble $\forall \theta \in \mathcal{C}.$ Let $p_\mathcal{E}$---the probability density function of the ensemble---be obtained as a mixture distribution where the mixture is given as  $ p_\mathcal{E} = \mathbb{E} p_{\theta}.$
Define $\mu_{\theta}$ and $\sigma_{\theta}^2$ as the mean and variance of each element in the ensemble, respectively. Then, the mean of the mixture is $\mu_{\mathcal{E}} := \mathbb{E}[\mu_{\theta}]$, and the variance~\cite{rudary2009predictive} is
\begin{equation} \label{eq:variance-mixture-regression_prop}
\begin{split}
\sigma_{\mathcal{E}}^2 &:= \mathbb{V}[p_\mathcal{E}] = \underbrace{\mathbb{E} [\sigma_{\theta}^2]}_{ \text{Aleatoric Uncertainty} } + \underbrace{ \mathbb{V}[\mu_{\theta}]}_{\text{Epistemic Uncertainty} }, \\
\end{split}
\end{equation}
where $\mathbb{E}$ refers to the expected value and $\mathbb{V}$ refers to the variance. Equation \eqref{eq:variance-mixture-regression_prop} formally provides the decomposition of the overall uncertainty of the ensemble into its individual components such that $ \mathbb{E} [\sigma_{\theta}^2]$ marginalizes the effect of $\theta$ and captures the aleatoric uncertainty and $\mathbb{V}[\mu_{\theta}]$ captures the spread of the prediction across different models and neglects the noise of the data, therefore capturing the epistemic uncertainty.

The overall estimate of the mean and the variance may then be written as
\begin{equation}
\begin{split}
\mu_{\mathcal{E}}     &  = \frac{1}{K}\sum_{\theta \in \mathcal{E} } \mu_{\theta} \\
\sigma_\mathcal{E}^2  &  = \underbrace{\frac{1}{K} \sum_{\theta \in \mathcal{E}} \sigma^2_\theta}_{ \text{Aleatoric Uncertainty} } + \underbrace{\frac{1}{K-1} \sum_{\theta \in \mathcal{E}} (\mu_\theta - \mu_\mathcal{E})^2}_{ \text{Epistemic Uncertainty}},
\label{eq:var-decomposition-2}
\end{split}
\end{equation}
where $K$ is the size of the ensemble. The total uncertainty here quantified by $\sigma_\mathcal{E}^2 $ is a combination of aleatoric and epistemic uncertainty.

Let us assume that our total data $\mathcal{D}$ is partitioned in the usual train-validation-test split manner as $\mathcal{D} = \mathcal{D}^{train} \cup \mathcal{D}^{valid} \cup \mathcal{D}^{test}$. A neural architecture configuration $\theta_a$ is a vector from the neural architecture search space $\Theta_a$, defined by a set of neural architecture decision variables. A hyperparameter configuration $\theta_h$ is a vector from the training hyperparameter search space $\Theta_h$ defined by a set of hyperparameters used for training (e.g., learning rate, batch size). The problem of joint neural architecture and hyperparameter search can be formulated as the following bilevel optimization problem:  
\begin{equation}
    \begin{aligned}
        &\theta_a^*, \theta_h^* = \argmax_{\theta_a,\theta_h} \frac{1}{N^{valid}}\sum_{\mathbf{x}, y \in \mathcal{D}^{valid}}\ell(\mathbf{x}, y;\theta_a,\theta_h,\theta_w^*) \\
        & \text{s.t. }  \theta_w^* =  \argmax_{\theta_w} \frac{1}{N^{train}}\sum_{\mathbf{x}, y \in \mathcal{D}^{train}} \ell(\mathbf{x}, y; \theta_a, \theta_h, \theta_w),
    \end{aligned}
    \label{eqn:pbopti3}
\end{equation}
where the best architecture decision variables $\theta_a^*$ and training hyperparameters values  $\theta_h^*$ are selected based on $\mathcal{D} = \mathcal{D}^{valid}$ and the corresponding weights $\theta_w$ are selected based on $\mathcal{D} = \mathcal{D}^{train}$.

\begin{algorithm2e}[!ht]
\footnotesize
\DontPrintSemicolon
\SetInd{0.5em}{0.5em}
\SetAlgoLined
\SetKwInOut{Input}{inputs}\SetKwInOut{Output}{output}
\SetKwFunction{RandomPoint}{random\_sample}
\SetKwFunction{SubmitEval}{submit\_for\_training}
\SetKwFunction{GetFinishedEval}{check\_finished\_training}
\SetKwFunction{Push}{push}
\SetKwFunction{EmptyList}{EmptyList}
\SetKwFunction{RandomSample}{random\_sample}
\SetKwFunction{SelectParent}{select\_parent}
\SetKwFunction{Mutate}{mutate}
\SetKwFunction{Tell}{tell}
\SetKwFunction{Ask}{ask}

\SetKwFor{For}{for}{do}{end}
\Input{P: population size, S: sample size, W: workers}
\Output{$\mathcal{E}$: ensemble of models}
    {\color{orange} \tcc{Initialization for AgEBO}}
    $population \leftarrow$ create\_queue($P$) \tcp{Alloc empty Q of size P}
    $BO \leftarrow$ Bayesian\_Optimizer()\\
    \For{$i\leftarrow 1$ \KwTo $W$}{
        $config.\theta_a \leftarrow$ \RandomPoint{$\Theta_a$}\\
        $config.\theta_h \leftarrow$ \RandomPoint{$\Theta_h$}\\
        \SubmitEval{config} \tcp{Nonblocking}So 
    }
    
    {\color{orange} \tcc{Optimization loop for AgEBO}}
    \While{stopping criterion not met}{
        \tcp{Query results}
        $results \leftarrow$ \GetFinishedEval()\\
        $\mathcal{C} \leftarrow \mathcal{C} \cup results$ \tcp{Add to catalogue population}
        \If{$|results| > 0$}{
            $population.$\Push{results} \tcp{Aging population}
            \tcp{Generate hyperparameter configs}
            $BO.$\Tell{$results.\theta_h, results.valid\_score$}\\
            $next \leftarrow$ $BO.$\Ask{$|$results$|$} \tcp{Generate architecture configs}
            \For{$i\leftarrow 1$ \KwTo $|results|$}{
                \eIf{$|population| = P$}{
                    $parent.config \leftarrow$ \SelectParent{population,S}\\
                    $child.config.\theta_a \leftarrow$ \Mutate{$parent.\theta_a$}
                }
                {
                $child.config.\theta_a \leftarrow$ \RandomPoint{$\Theta_a$}
                }
                $child.config.\theta_h \leftarrow next[i].\theta_h$ \\
                \SubmitEval{$child.config$} \tcp{Nonblocking}
            }
        }
    }
    {\color{orange} \tcc{Ensemble selection}}
    $\mathcal{E} \leftarrow \{\}$\\
    \While{$|\mathcal{E}.unique()| \leq K$}{
        $\theta^* \leftarrow \argmin_{\theta \in \mathcal{C}} \ell(\theta,X,y)$ \tcp{Find lowest validation loss} 
        $\mathcal{E} \leftarrow \mathcal{E} \cup \{ \theta^* \}$ \tcp{Append model to ensemble}
    }
    return $\mathcal{E}$
 \caption{AutoDEUQ: Neural architecture and hyperparameter search for ensemble construction.}
 \label{alg:AgEBO}
\end{algorithm2e}

The pseudo-code of the AutoDEUQ is shown in Algorithm \ref{alg:AgEBO}. To perform a joint neural architecture and hyperparameter search, we leverage aging evolution with asynchronous Bayesian optimization (AgEBO)~\cite{egele2020agebo}. 

Aging evolution (AgE)~\cite{real2019regularized} is a parallel neural architecture search (NAS) method for searching over the architecture space. 
The AgEBO method follows the manager-worker paradigm, wherein a manager node runs a search method to generate multiple NNs and $W$ workers (compute nodes) train them simultaneously. The AgEBO method constructs the initial population by sampling $W$ architecture and  $W$ hyperparameter configurations and concatenating them (lines~1--7). The NNs obtained by using these concatenated configurations are sent for simultaneous  evaluation on $W$ workers (line~6). The iterative part (lines~8--26) of the method checks whether any of the workers finish their evaluation (line~9), collects validation metric values from the finished workers, and uses them to generate the next set of architecture and hyperparameter configurations for simultaneous evaluation to fill up the free workers that finished their evaluations (lines~11--25). 
At a given iteration, in order to generate a NN, architecture, and hyperparameter configurations are generated in the following way. From the incumbent population, $S$ NNs are sampled (line~17). A random mutation is applied to the best of $S$ NNs to generate a child architecture configuration (line~18). This mutation is obtained by first randomly selecting an architecture decision variable from the selected NN and replacing its value with another randomly selected value excluding the current value. The new child replaces the oldest member of the population. The AgEBO optimizes the hyperparameters ($\theta_h$) by marginalizing the architecture decision variables ($\theta_a$). At a given iteration, to generate a hyperparameter configuration, the AgEBO uses a (supervised learning) model $M$ to predict a point estimate (mean value) $\mu(\theta_h^i)$ and standard deviation $\sigma(\theta_h^i)$ for a large number of unseen hyperparameter configurations. The best configuration is selected by ranking all sampled hyperparameter configurations using the upper confidence bound  acquisition function, which is parameterized by $\kappa \geq 0$ that controls the trade-off between exploration and exploitation. To generate multiple hyperparameter configurations at the same time, the AgEBO leverages a multipoint acquisition function based on a constant liar strategy~\cite{hiot_kriging_2010}. 

The catalog $\mathcal{C}$ of NN models is obtained by running AgEBO and storing all the models from the runs. To build the ensemble  $\mathcal{E}$ of models from $\mathcal{C}$, we adopt a top-$K$ strategy (lines~27--32). Here, the idea is to select the $K$ highest-performing models for the validation data and construct the ensemble from its individual predictions. This approach is appropriate when the validation data is representative of the generalization task (i.e., big enough, diverse enough, with good coverage)~\cite{caruana_ensemble_2004}. 

The architecture search space is modeled by using a directed acyclic graph, which starts and ends with input and output nodes, respectively. They represent the input and output layers of NN, respectively. Between the two are intermediate nodes defined by a series of variable $\mathcal{N}$ and skip connection $\mathcal{SC}$ nodes. Both types of nodes correspond to categorical decision variables.
The variable nodes model dense layers with a list of different layer configurations. The skip connection node creates a skip connection between the variable nodes. This second type of node can take two values: disable or create the skip connection.
For a given pair of consecutive variable nodes $\mathcal{N}_{k}$, $\mathcal{N}_{k+1}$, three skip connection nodes $\mathcal{SC}^{k+1}_{k-3}, \mathcal{SC}^{k+1}_{k-2}, \mathcal{SC}^{k+1}_{k-1}$  are created. These  nodes allow for connection to the previous nonconsecutive variable nodes $\mathcal{N}_{k-3}, \mathcal{N}_{k-2}, \mathcal{N}_{k-1}$, respectively. 


For the hyperparameter search space, we use a learning rate in the continuous range $[10^{-4},10^{-1}]$ with a log-uniform prior;  a batch size in the discrete range $[32,\ldots,b_{max}]$ (where $b_{max}=256$) with a log-uniform prior; an optimizer in 
$\{
    \def\OldComma{,}
    \catcode`\,=13
    \def,{%
      \ifmmode%
        \OldComma\discretionary{}{}{}%
      \else%
        \OldComma%
      \fi%
    }%
\text{sgd}, \text{rmsprop}, \text{adagrad}, \text{adam} 
\}$; 
a patience value of 15 for the reduction of the learning rate, and value of 20 early stopping of training. The individual definitions of the architecture search space are experiment specific and are explored in the respective problem definitions below. Models are checkpointed during their evaluation based on the minimum validation loss achieved.
We note that AutoDEUQ is deployed in the DeepHyper package, designed for scalable automatic machine learning on leadership class high-performance computers \cite{balaprakash2018deephyper}.

\subsection{Dataset}

In this study we use the open-source National Oceanic and Atmospheric Administration (NOAA) Optimum Interpolation sea surface temperature V2 data set (henceforth NOAA-SST).\footnote{Available at https://www.esrl.noaa.gov/psd/}  This data set has a strong periodic structure due to seasonal fluctuations in addition to rich fine-scaled phenomena due to complex ocean dynamics. Weekly-averaged NOAA-SST snapshots are available on a quarter-degree grid which is sub-sampled to a one-degree grid for the purpose of demonstrating our proposed methodology in a computationally efficient manner. This data set, at the 1-degree resolution, has previously been used in several data-driven analysis tasks (for instance, see \cite{kutz2016multiresolution,callaham2019robust} for specific examples), particularly from the point of view of extracting seasonal and long-term trends as well as for flow-field recovery \citep{maulik2020probabilistic}. Each ``snapshot'' of data originally corresponds to an array of size 360 $\times$ 180 (i.e., arranged according to the longitudes and latitudes of a one-degree resolution). 
However, for effective utilization in forecasting and flow-reconstruction, a mask is used to remove missing locations in the array that corresponds to the land area. Furthermore, it should be noted that regressions are performed for those coordinates which correspond to oceanic regions alone, and inland bodies of water are ignored. The non-zero data points then are subsequently flattened to obtain a column vector for each snapshot of our training and test data. We note that this data is available from October 22, 1981, to June 30, 2018 (i.e., 1,914 snapshots for the weekly averaged temperature). 

\subsection{Task 1: Forecasting}

For our forecasting task, we utilize the period of October 22, 1981, to December 31, 1989. The rest (i.e., 1990 to 2018) is used for testing. Our final number of snapshots for training amounts to 427, and for testing amounts to 1487. This train-test split of the data set is a common configuration for data-driven studies \cite{callaham2019robust} and the 8-year training period captures several short and long-term trends in the global sea surface temperature. Individual training samples are constructed by selecting a window of inputs (from the past) and a corresponding window of outputs (for the forecast task in the future) from the set of 427 training snapshots from the NOAA dataset. From the perspective of notation, if $\bm{y}_t$ is a snapshot of training data, $t= 1,2,\dots,427$, a forecasting technique may be devised by learning to predict $\bm{y}_{t+1}, \dots, \bm{y}_{t+\tau}$ given $\bm{y}_t, \bm{y}_{t-1}, \dots, \bm{y}_{t-\tau}$. We note that this forecast is performed non-autoregressively---that is, the data-driven method \emph{is not} utilized for predictions beyond the desired window size $\tau$. Therefore, it is always assumed that the \emph{true} $\bm{y}_t, \bm{y}_{t-1}, \dots, \bm{y}_{t-\tau}$ is available prior to making predictions. This means that given a window of true inputs (for example obtained via an observation of the state of the system given re-analysis data), a forecast is made for a series of outputs (corresponding to the window length) before a metric of accuracy is computed for optimization. This is in contrast to a situation where simply one step of a prediction is utilized for computing a fitness metric which adversely affects the ability of the predictive method for longer forecasts into the future (due to amplifying errors with each forecast step). The window size for the set of experiments on this data set is fixed at 8 weeks. The window-in and window-out construction of the training data leads to a final training data set of size 411 samples. Since this data set is produced by combining local and satellite temperature observations, it represents an attractive forecasting task for \emph{non-intrusive} data-driven methods without requiring the physical modeling of underlying processes.

Subsequently, proper orthogonal decomposition (POD) provides a systematic method to project dynamics of a high-dimensional system onto a lower-dimensional subspace. We suppose that a single snapshot of the full system is a vector in $\mathbb{R}^N$, where $N$ could be the number of grid points at which the field is resolved. Observing the system across a number of time points gives us the snapshots $\bm{y}_1, \dots, \bm{y}_T$, with mean subtracted by convention. The aim of POD is to find a small set of orthonormal basis vectors $\bm{v_1}, \dots, \bm{v}_M$, with $M \ll N$, which approximates the spatial snapshots,
$\bm{y}_t \approx \sum_{j=1}^M a_j(t)\bm{v}_j, \quad t=1,\dots,T,$
and so allows us to approximate the evolution of the full $N$ dimensional system by considering only the evolution of the $M$ coefficients $a_j(t)$. POD chooses the basis, $\bm{v}_j$, to minimize the residual with respect to the $L_2$ norm, 
$
R = \sum_{t=1}^{T} || \bm{y}_t - \sum_{j=1}^M a_j(t) \bm{v}_j ||^2.
$
Defining the snapshot matrix, $\bm{S} = [ \bm{y}_1 | \cdots | \bm{y}_T ]$, the optimal basis is given by the $M$ eigenvectors of $\bm{SS}^T$, with largest eigenvalues, after which, the coefficients are found by orthogonal projection, $\mathbf{a}(t) = \langle \bm{y}_t,\bm{v}\rangle$~\cite{ModalAnalysisRev}. The coefficients $\mathbf{a}(t)$ correspond to a time-series with an $M-$dimensional state vector which is the focus of our temporal forecasting.

For our learning, we take only the training data snapshots, say $\bm{D}_1, \dots,\bm{D}_T$, from which we calculate the mean $\bar{\bm{D}} = (1/T) \sum_t \bm{D}_t$, hence defining the mean subtracted snapshots $\bm{y}_t = \bm{D}_t - \bar{\bm{D}}$. We then create the snapshot matrix, $\bm{S}$, and find numerically the $M$ eigenvectors of $\bm{S}\bm{S}^T$ with largest eigenvalues. From this, we train models to forecast the coefficients making predictions of future coefficients given previous ones. Note that our model is also capable of making predictions for the aleatoric uncertainty, i.e., we can predict the variance of the output data given a Gaussian likelihood assumption for the loss function. This implies that we also predict variances in addition to the deterministic mean, i.e., 
\small{
\begin{align}
f^\dagger ( \bm{a}(t), \bm{a}(t-1),\dots, \bm{a}(t-\tau)) &= 
((\hat{\bm{a}}(t+1),\sigma_{\hat{\bm{a}}}(t+1)), (\hat{\bm{a}}(t+2),\sigma_{\hat{\bm{a}}}(t+2)), \dots, (\hat{\bm{a}}(t+\tau),\sigma_{\hat{\bm{a}}}(t+\tau))) \\
(\bm{a}(t+1), \bm{a}(t+2), \dots, \bm{a}(t+\tau)) & \approx (\hat{\bm{a}}(t+1), \hat{\bm{a}}(t+2), \dots, \hat{\bm{a}}(t+\tau)) 
\end{align}
}

Here, $\tau$ corresponds to the time-delay that is embedded into the input feature space for the purpose of forecasting. In this article, $\tau$ also stands for the forecast length obtained from the fit model although in the general case, these quantities can be chosen to be different. We can now test for predictions on unseen data, $\bm{E}_1,\dots,\bm{E}_Q$, where $\bm{E}_t$ is an unseen snapshot of data at time $t$ and $Q$ is the total number of test snapshots. Note that these test snapshots are obtained for a time-interval that has not been utilized for constructing the POD basis vectors. We proceed by taking the mean $\bar{\bm{D}}$, and vectors $\bm{v}_j$ calculated from the training data to get test coefficients at an instant in time
$
a_j(t) = \langle \bm{E}_t - \bar{\bm{D}}, \bm{v}_j \rangle, \quad j = 1,\dots,M,
$
which will be used with the model $f^\dagger$ to make future predictions in the unseen testing interval. The prediction for the coefficients $\hat{\bm{a}}$, can be converted into predictions in the physical space by taking $\bar{\bm{D}} + \sum_j \hat{a}_j \bm{v}_j$. This procedure only makes use of testing data to pass into the model, not to train the model in any way. Crucially, to make a forecast of $\bm{E}_{t+1}, \bm{E}_{t+2}, \dots, \bm{E}_{t+\tau}$, only previous measurements $\bm{E}_{t},\bm{E}_{t-1}, \dots, \bm{E}_{t-\tau}$ are needed. The connection to AgeBO comes in through how one may obtain several models like $f^{\dagger}$ which may subsequently be used for forecasting.

\subsection{Task 2: Field reconstruction from sparse sensors}

In this section, we consider the problem of reconstructing the sea-surface temperature data given sparse observations of the flow-field. We use 20 years of data (1040 snapshots spanning 1981 to 2001) as the training data set, while the test data set is prepared from 874 snapshots spanning from year 2001 to 2018.
Thus the regression problem is given by attempting to find a map
\begin{align}
    \bm{y}_t \approx g^\dagger ( H[\bm{y}_t] )
\end{align}
where $H$ is the observation operator mapping the full state to sparse observations. We note that unlike the previous forecasting problem, there is no requirement of a normalization procedure on the data for constructing our data-driven models. This test setting represents extrapolation in time but not physics, since the data set has the influence of seasonal periodicity. The aforementioned problem setting follows the work of \cite{CMB2019} who attempted to reconstruct fluid-flow fields from local sensors using sparse representations. As in their article, the input sensors for the baseline model are chosen randomly from the region of $50{^\circ}$ S to $50{^\circ}$ N. We utilize the same configuration as our previous articles \citep{maulik2020neural} that have also used this benchmark problem for algorithmic development for the purpose of consistency. 

\section{Results}

In the following section, we detail our results for the forecasting and reconstruction experiments.

\subsection{Forecasting}

\begin{figure}
    \centering
    \includegraphics[width=0.5\textwidth]{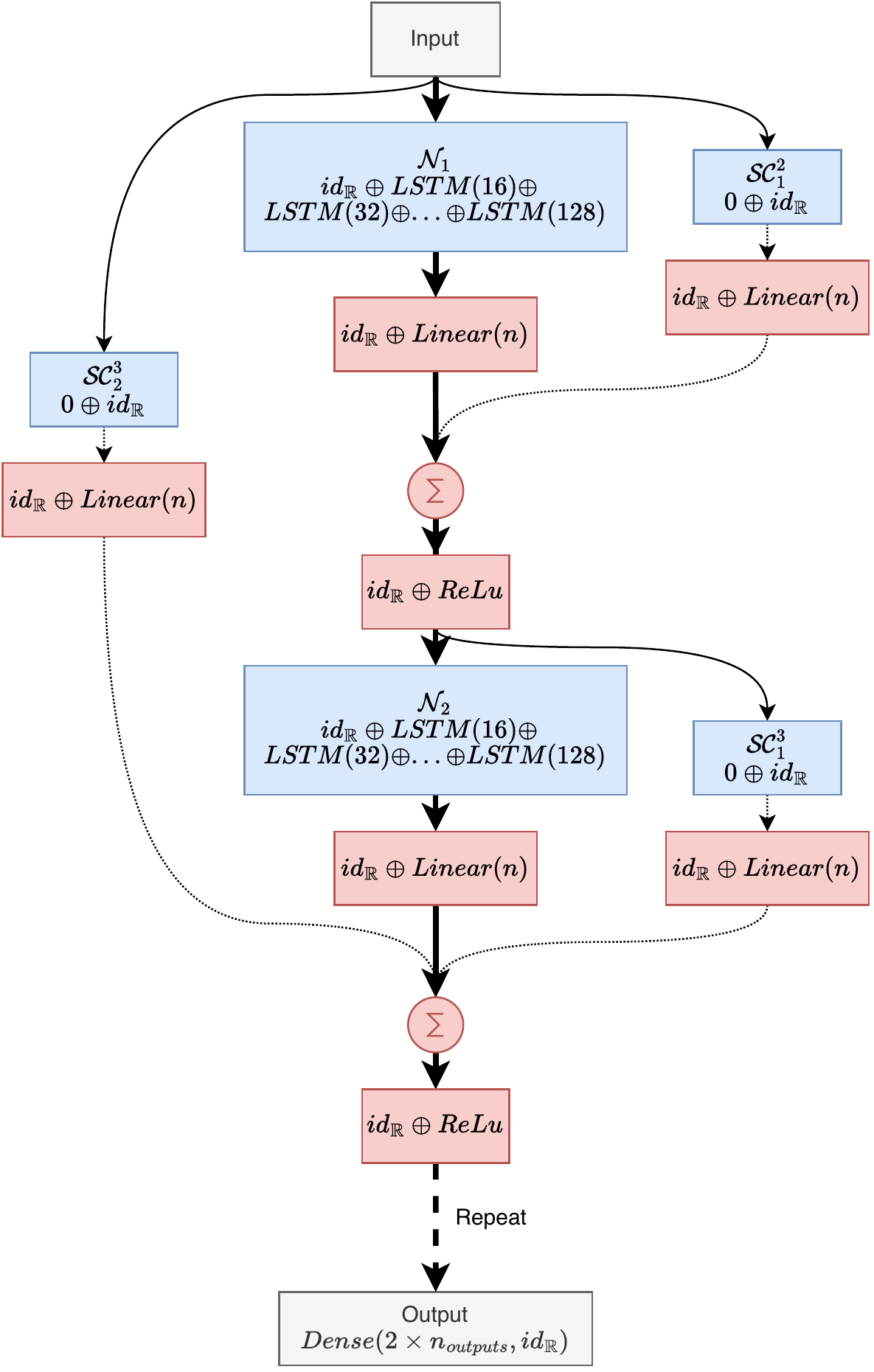}
    \caption{A description of the neural architecture search space constructed for the flow-forecasting task.}
    \label{fig_space_lst}
\end{figure}

We first assess the capability of DeepHyper NAS for discovering high-performing models that forecasting the NOAA sea-surface temperature. Here, forecasts are performed for 8-week output windows, given 8-week input information. The inputs and outputs of our model are given by sets of POD coefficients. We devise a search-space of stacked long short-term memory (LSTM) neural network models as shown in Figure \ref{fig_space_lst}. In this search space, we utilize five fully-connected (i.e., stacked) LSTM cells that are then allowed to have interactions between them using skip connections. The output of this neural architecture is a sequence of coefficients (with corresponding mean and aleatoric variance) in the latent space of the dynamics. 

\begin{figure}
    \centering
    \mbox{
    \subfigure[Convergence]{\includegraphics[width=0.48\textwidth]{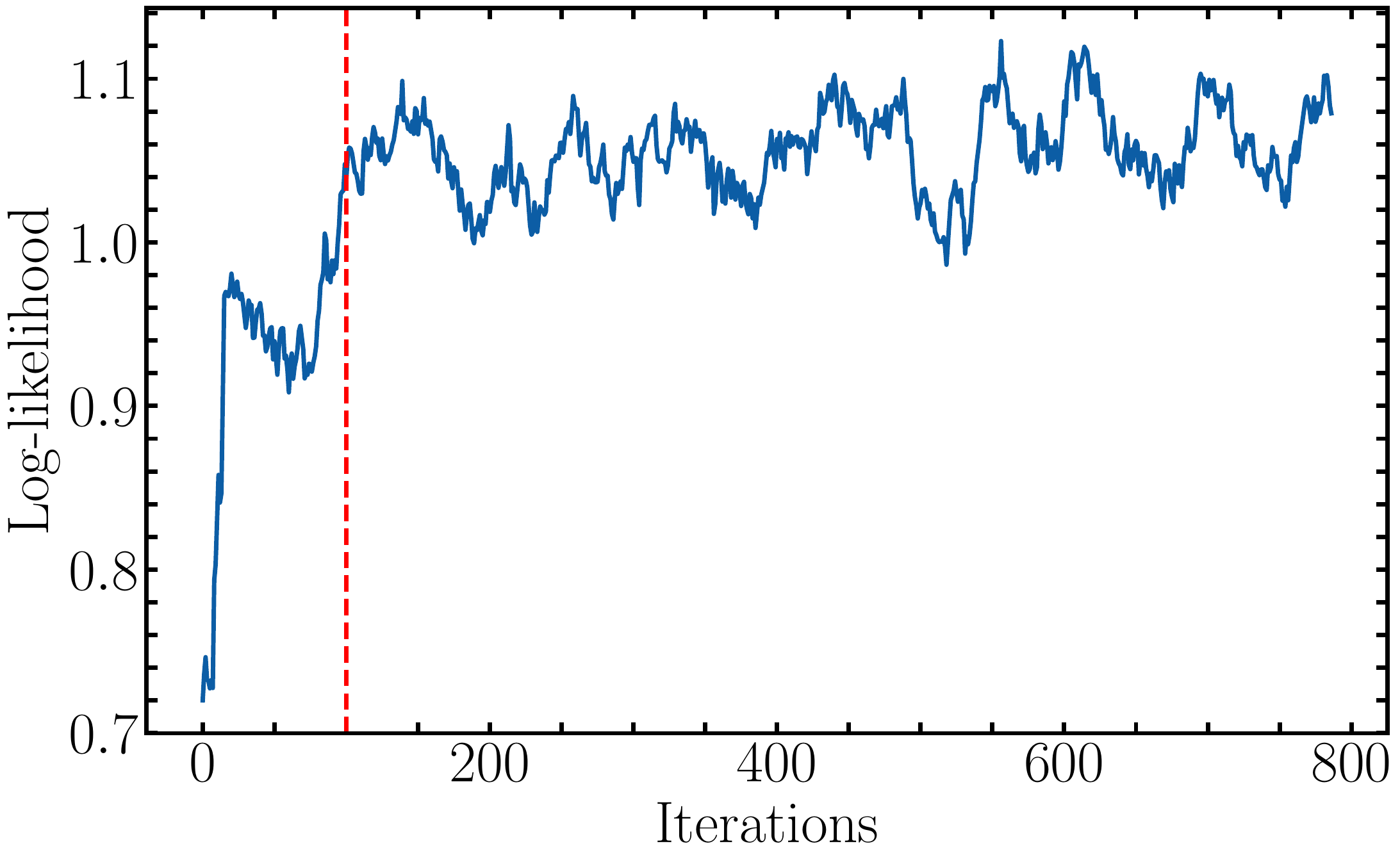}}
    \subfigure[Model spectrum]{\includegraphics[width=0.5\textwidth]{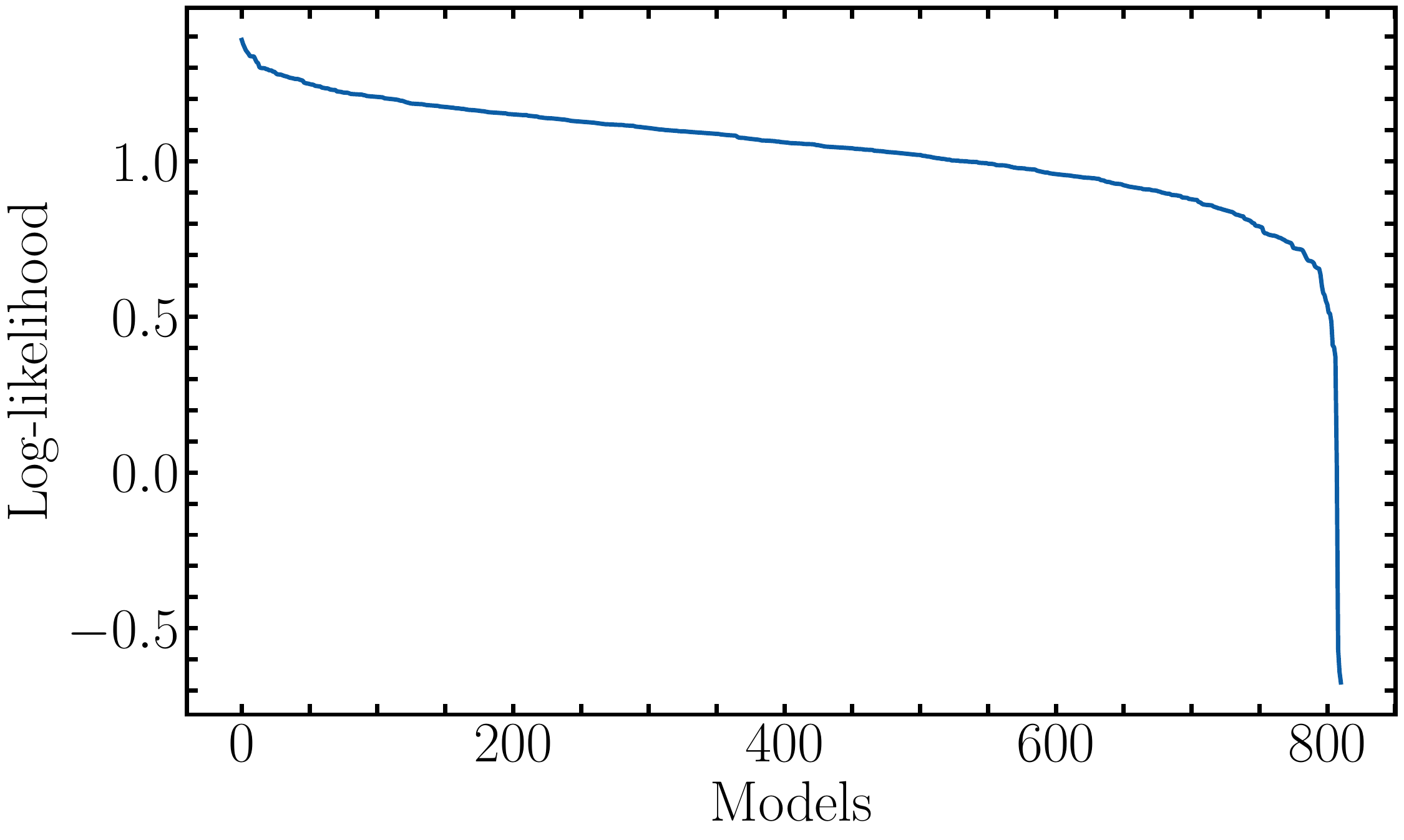}}
    }
    \caption{Convergence of neural architecture search shown through moving averages (left) and spectrum of model performance (right) for the forecasting task. The red dashed line indicates the start of intelligent search space exploration.}
    \label{nas_stats_lstm}
\end{figure}

In Figure \ref{nas_stats_lstm}, we show results from the NAS of stacked LSTM architectures. We first show that the search for neural architectures demonstrates improving performance with increasing number of iterations of the outer-loop differential evolution algorithm with a sharp increase in the log-likelihood over the first few iterations followed by a gradual convergence. In addition, we also show the performance of various models obtained at the final iteration, where a large majority of the models have superior performance in comparison to a few that are abnormally accurate or inaccurate.

\begin{figure}
    \centering
    \mbox{
    \subfigure[Week 1]{\includegraphics[width=0.4\textwidth]{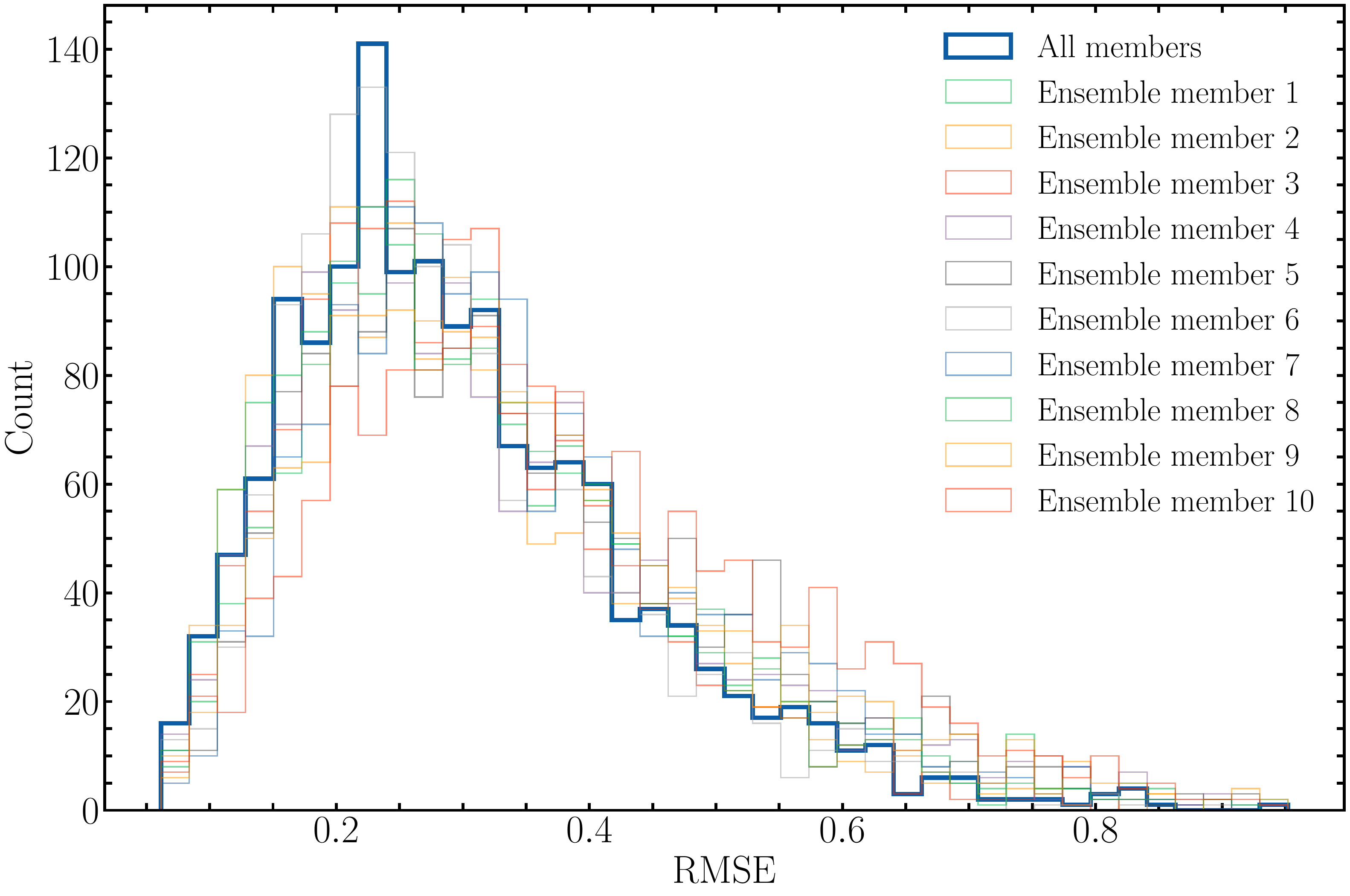}}
    \subfigure[Week 2]{\includegraphics[width=0.4\textwidth]{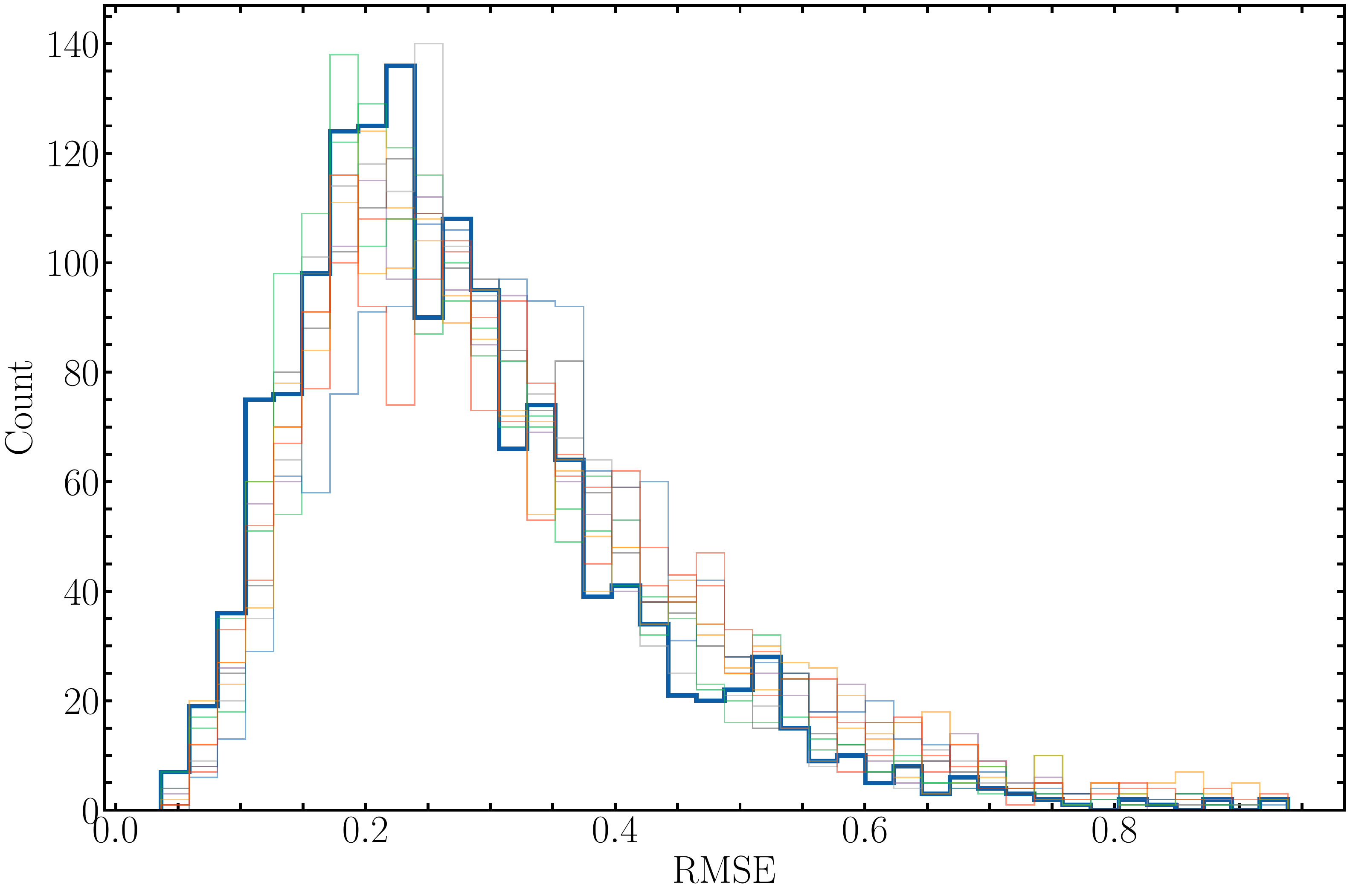}}
    } \\
    \mbox{
    \subfigure[Week 3]{\includegraphics[width=0.4\textwidth]{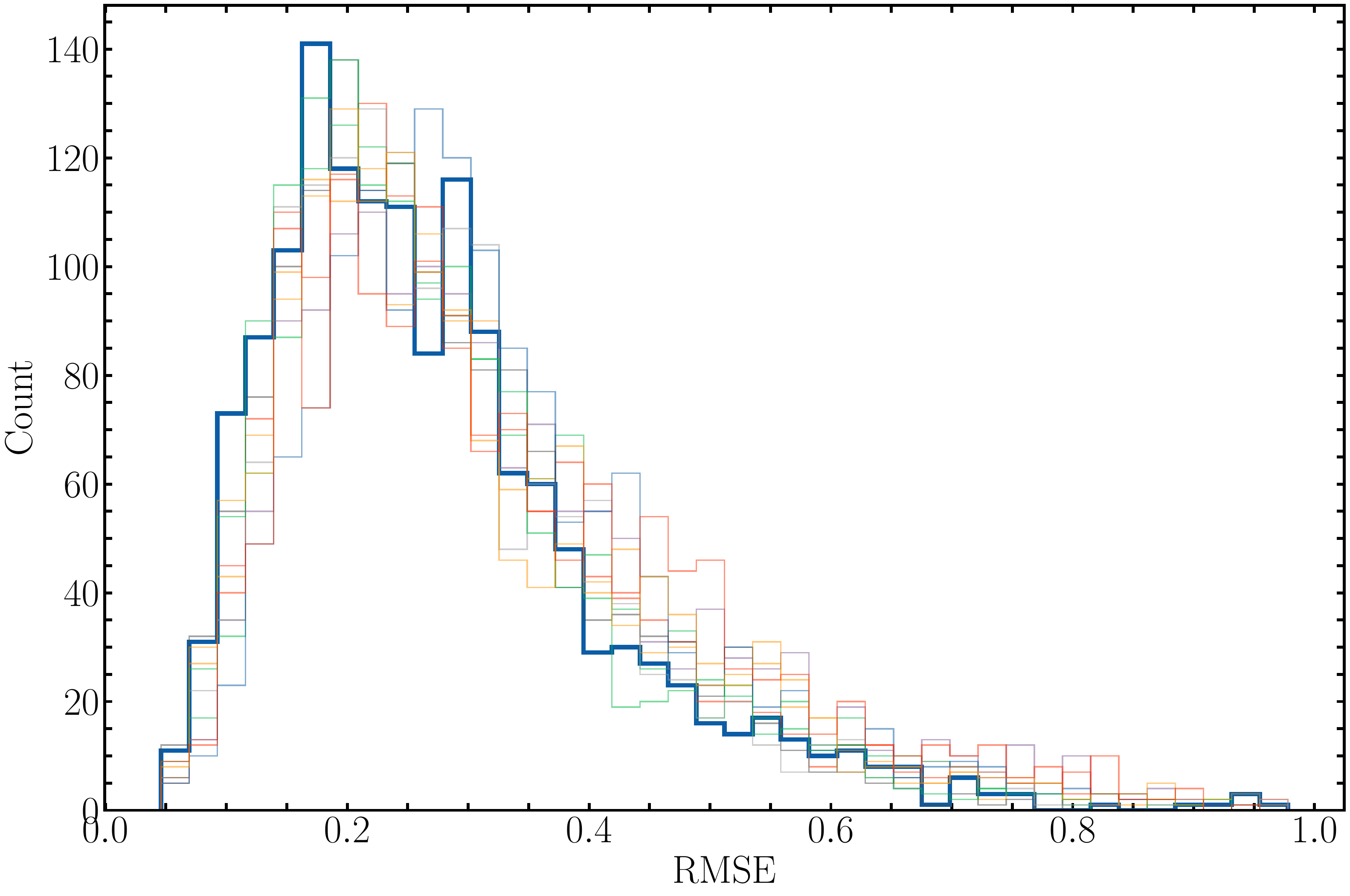}}
    \subfigure[Week 4]{\includegraphics[width=0.4\textwidth]{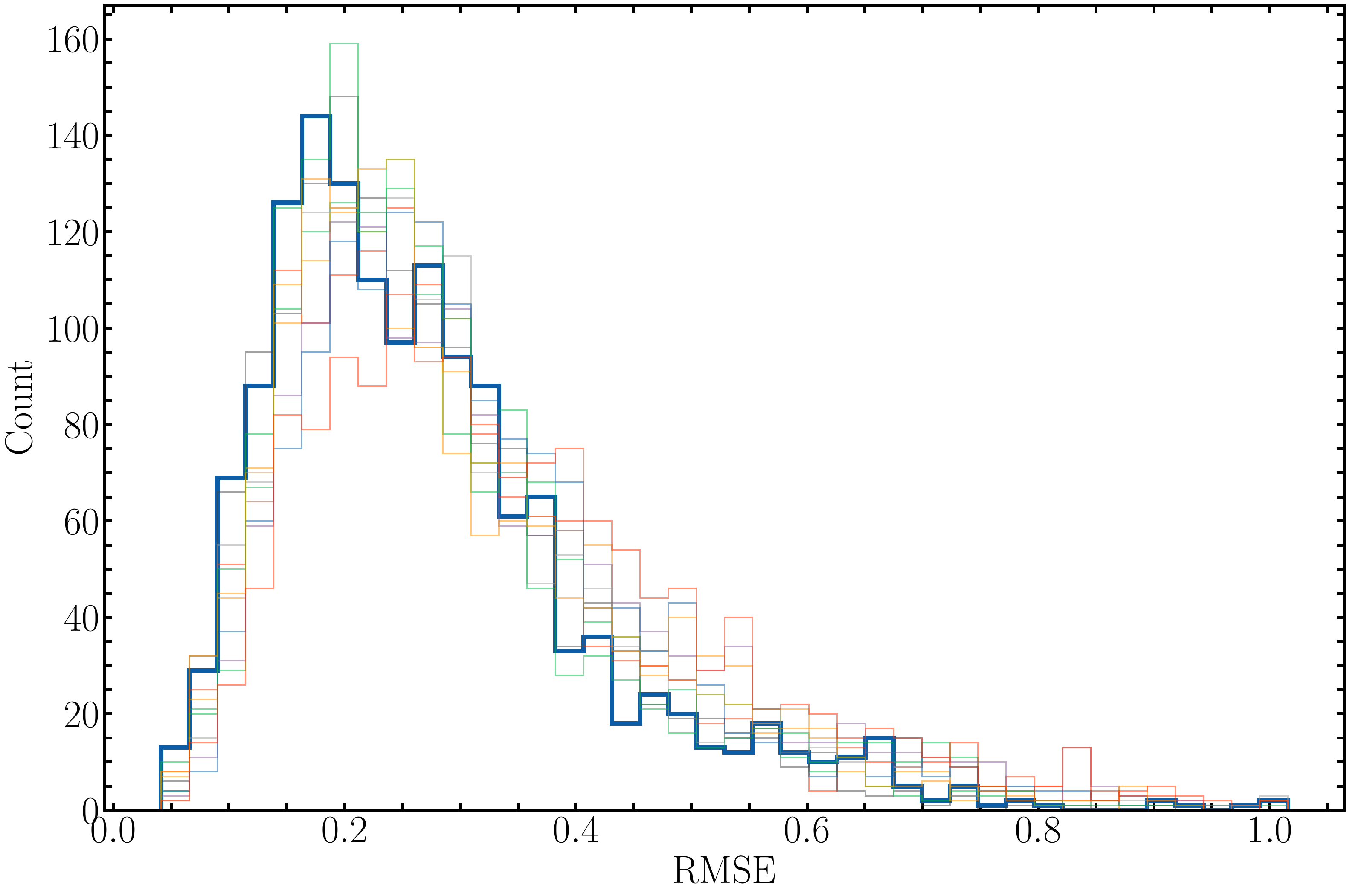}}
    } \\
    \mbox{
    \subfigure[Week 5]{\includegraphics[width=0.4\textwidth]{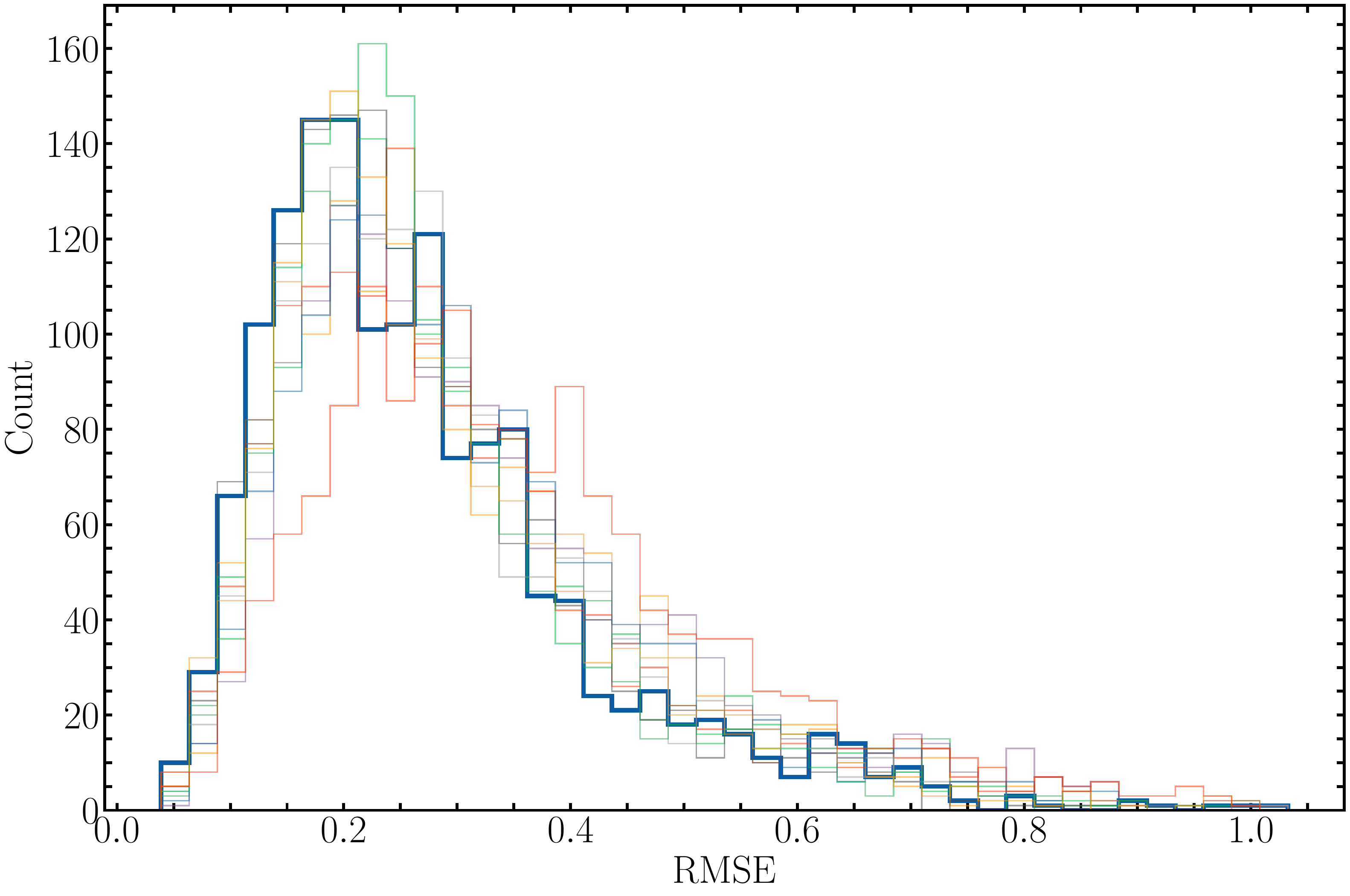}}
    \subfigure[Week 6]{\includegraphics[width=0.4\textwidth]{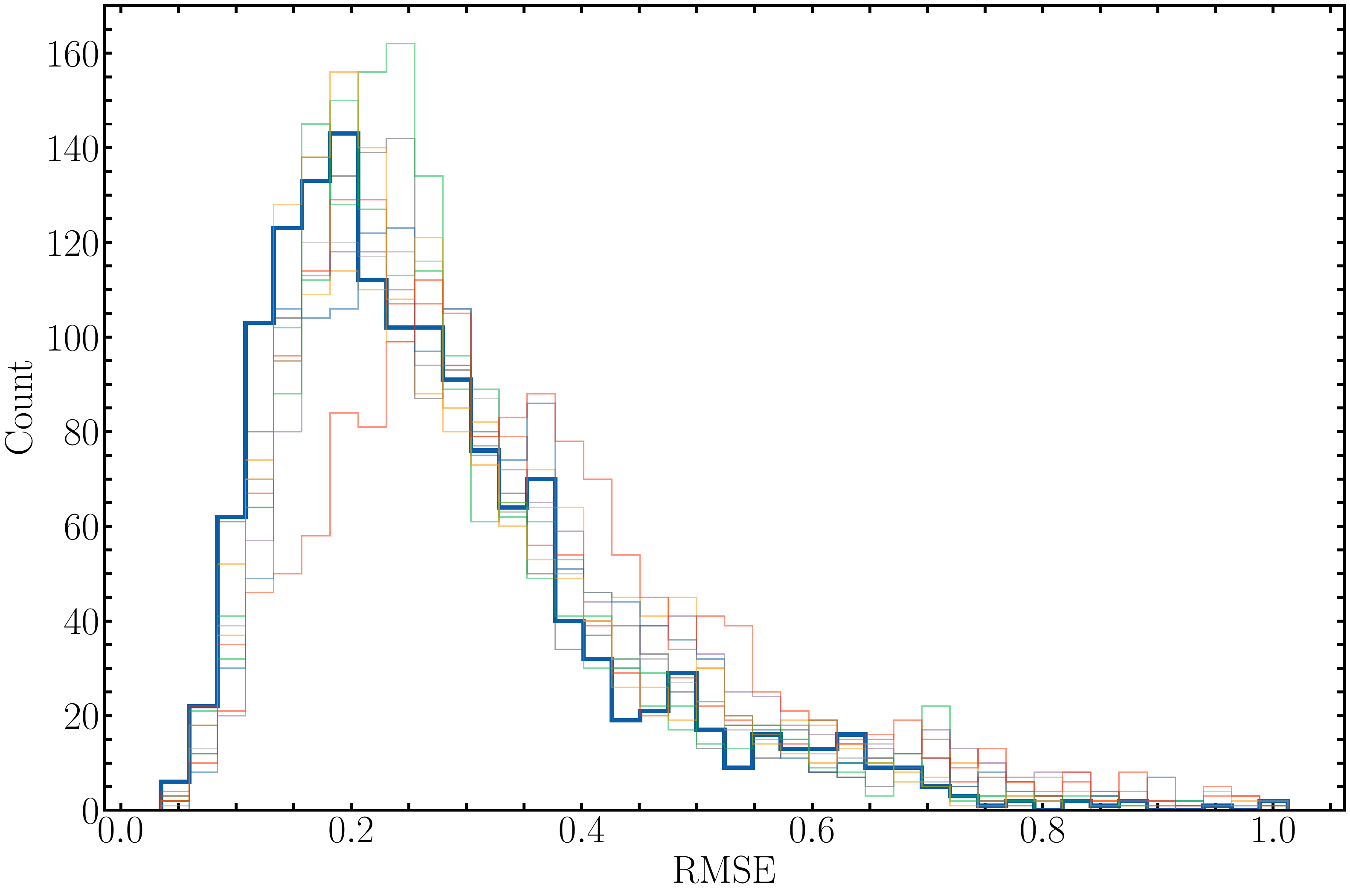}}
    } \\
    \mbox{
    \subfigure[Week 7]{\includegraphics[width=0.4\textwidth]{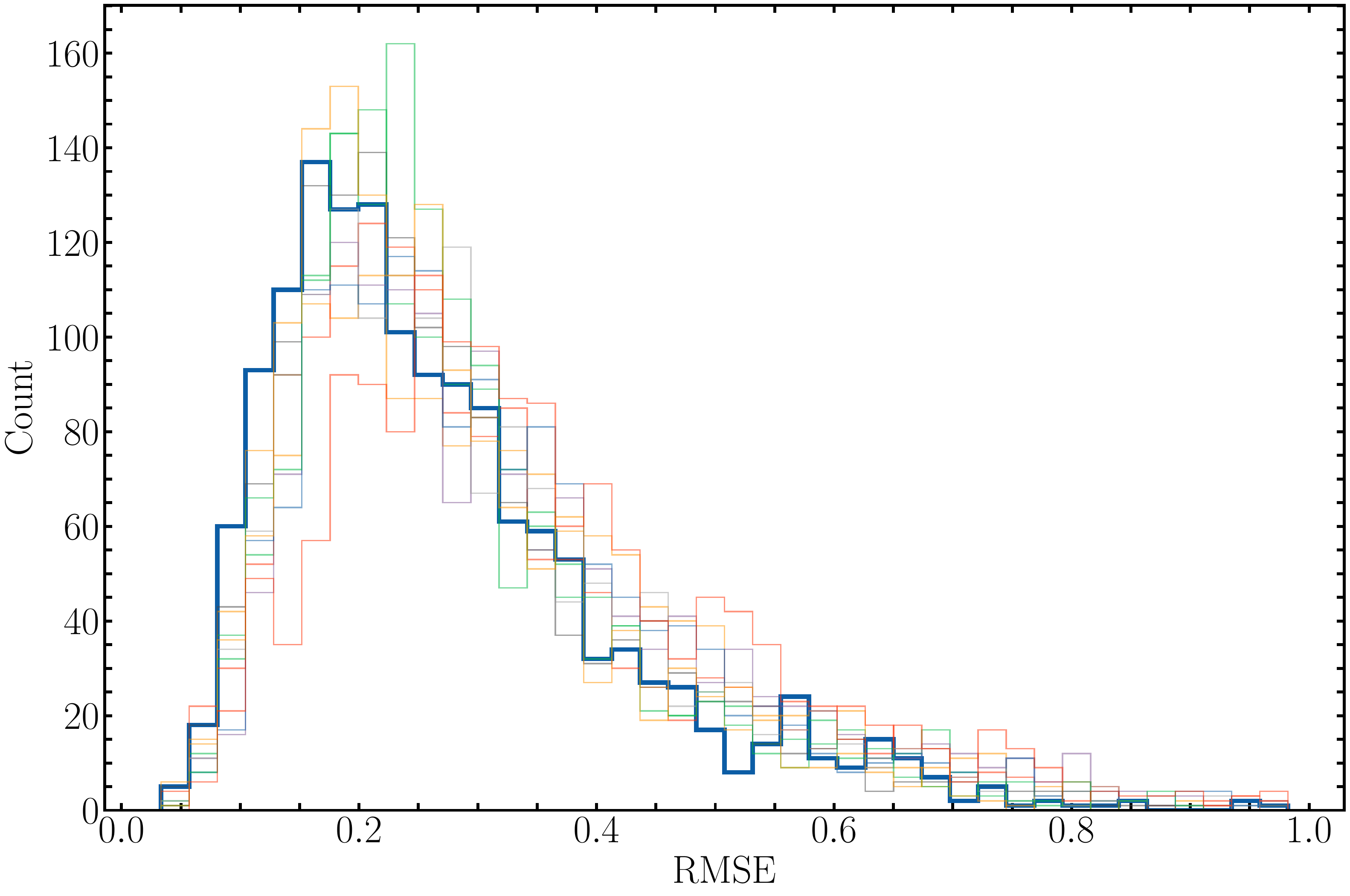}}
    \subfigure[Week 8]{\includegraphics[width=0.4\textwidth]{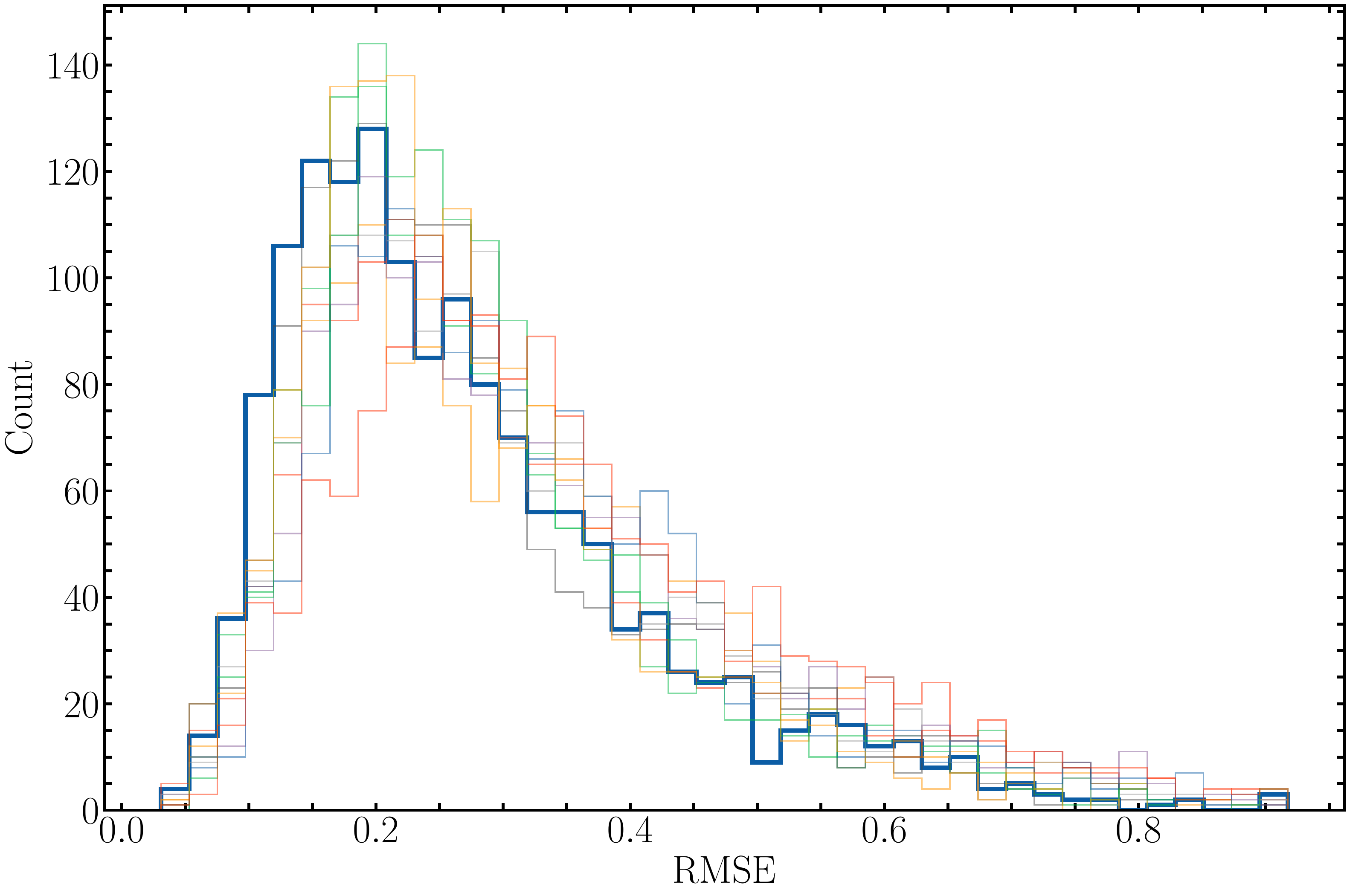}}
    }
    \caption{Pointwise RMSE histograms for various forecast weeks showing individual member bin counts as well as ensemble mean performance. Most points on the grid show a RMSE of around 0.2.}
    \label{nas_ensemble_rmse_hist}
\end{figure}

We extract 10 neural architectures from the entire search population and perform ensemble analyses with them. Note that the choice of the number of architectures is a user-defined parameter. We first compute the root-mean-squared-error of each neural architecture with respect to the truth for the testing range of the data set. We remind the reader that the output of a neural architecture member of the ensemble is a mean value for a Gaussian as well as a variance that represents the aleatoric uncertainty. We compute error statistics for individual ensemble members solely with the mean value predicted at the output layer. For computing the prediction of the entire ensemble, we compute a mean of the predictions from each member of the ensemble and then compute corresponding error statistics. We plot histograms of the RMSE for various weeks in Figure \ref{nas_ensemble_rmse_hist} which demonstrate how ensemble statistics reduce the potential heavy-tailed errors of individual ensemble members. 

\begin{figure}
    \centering
    \mbox{
    \subfigure[Week 1]{\includegraphics[width=0.4\textwidth]{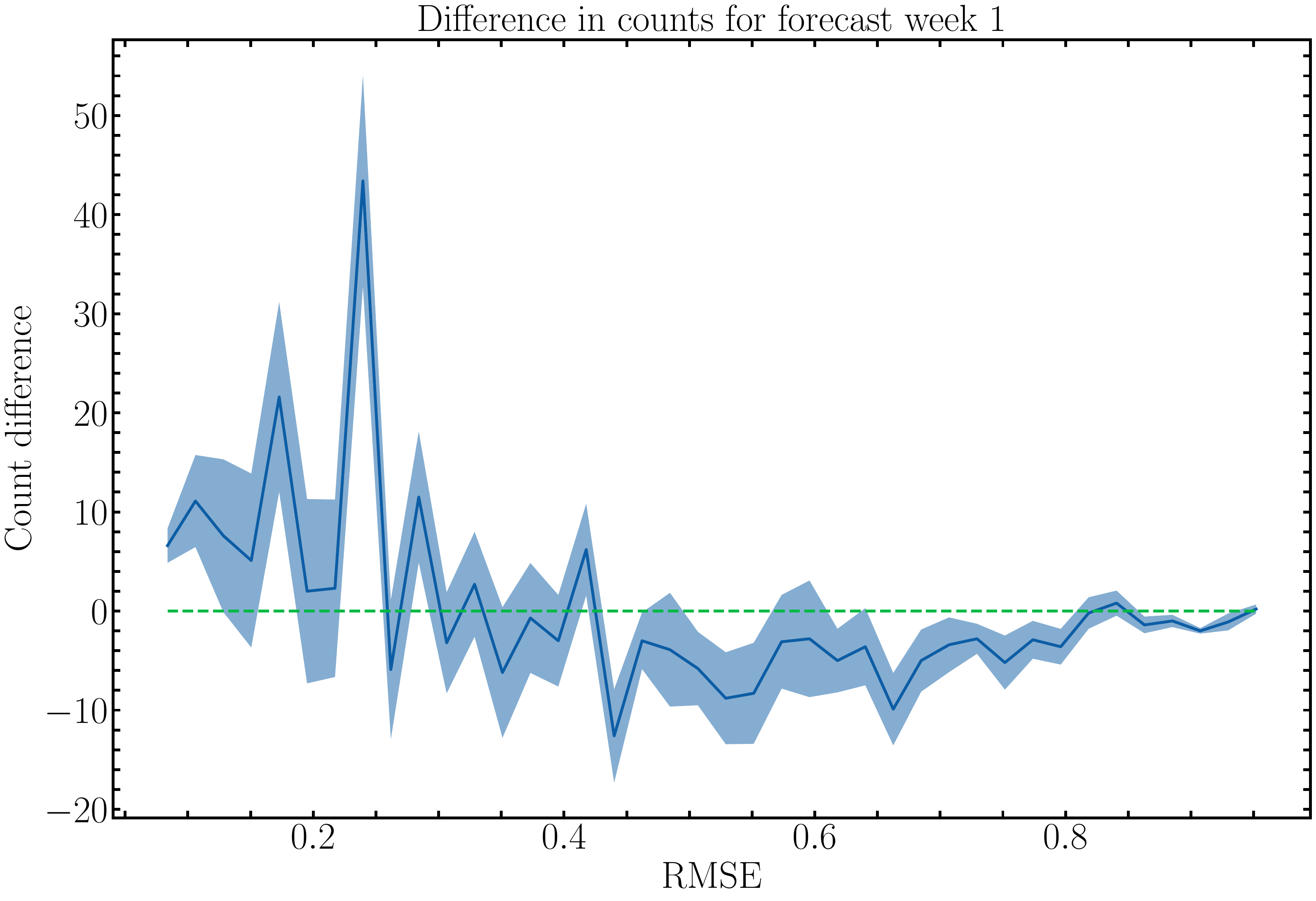}}
    \subfigure[Week 2]{\includegraphics[width=0.4\textwidth]{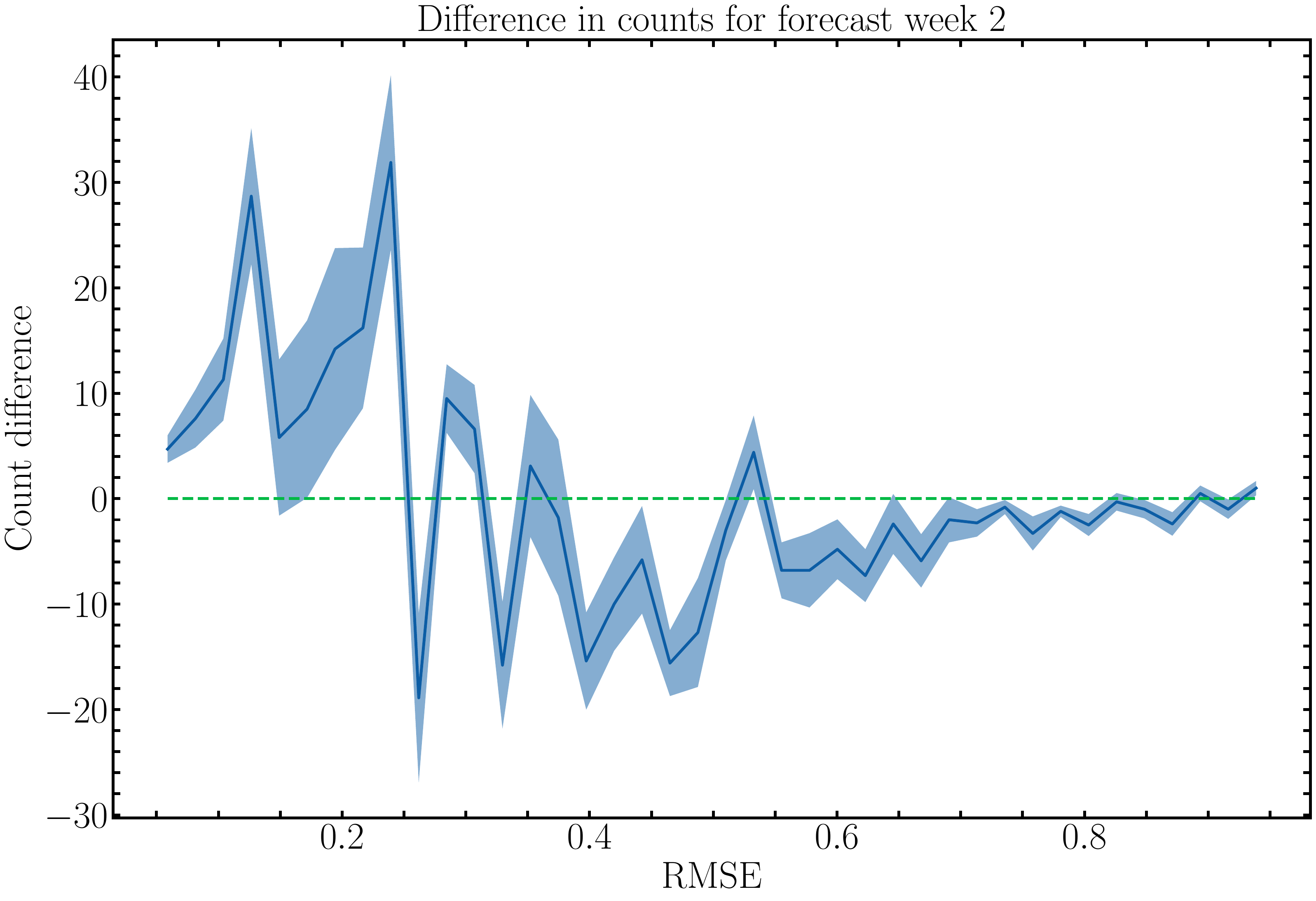}}
    } \\
    \mbox{
    \subfigure[Week 3]{\includegraphics[width=0.4\textwidth]{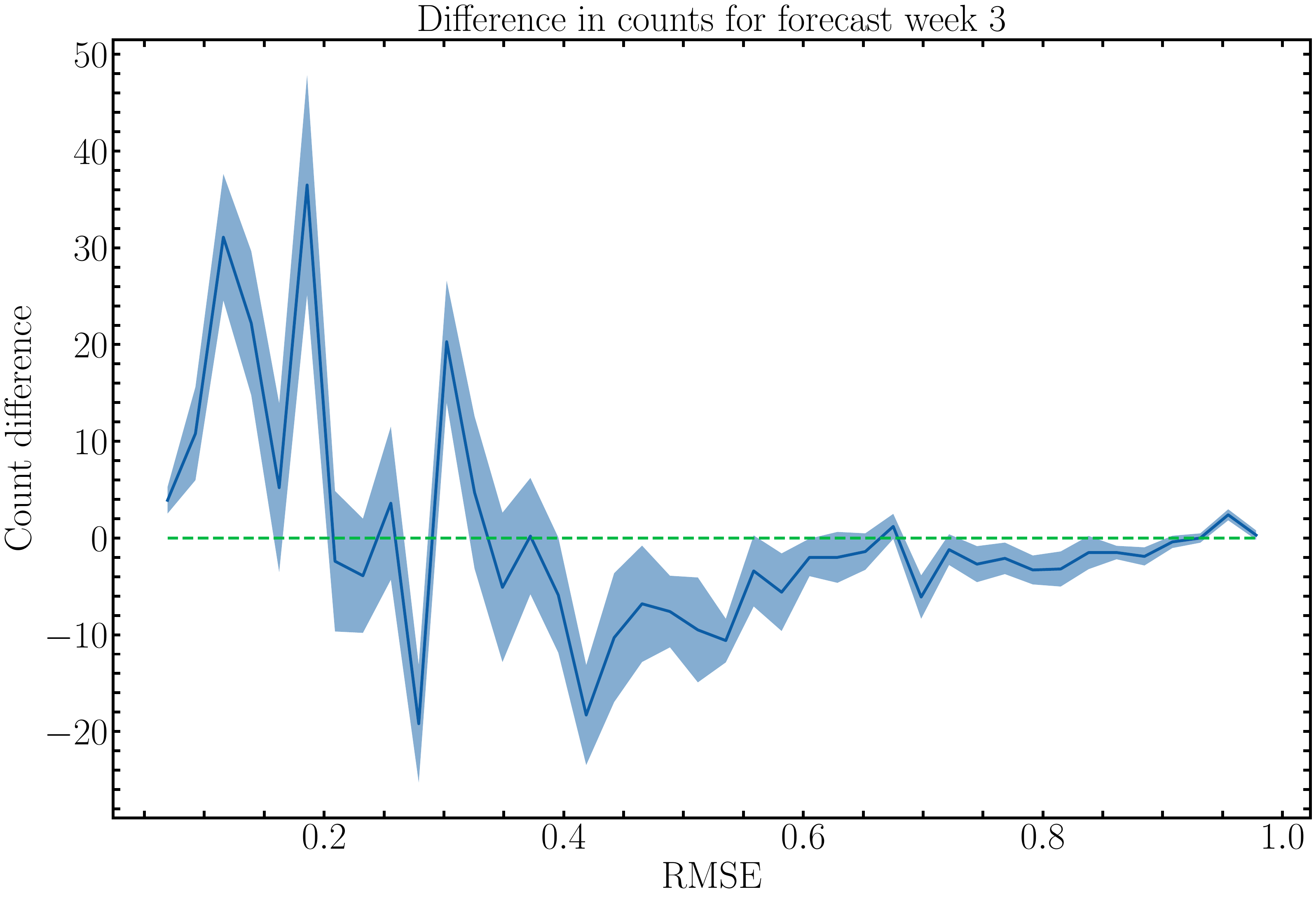}}
    \subfigure[Week 4]{\includegraphics[width=0.4\textwidth]{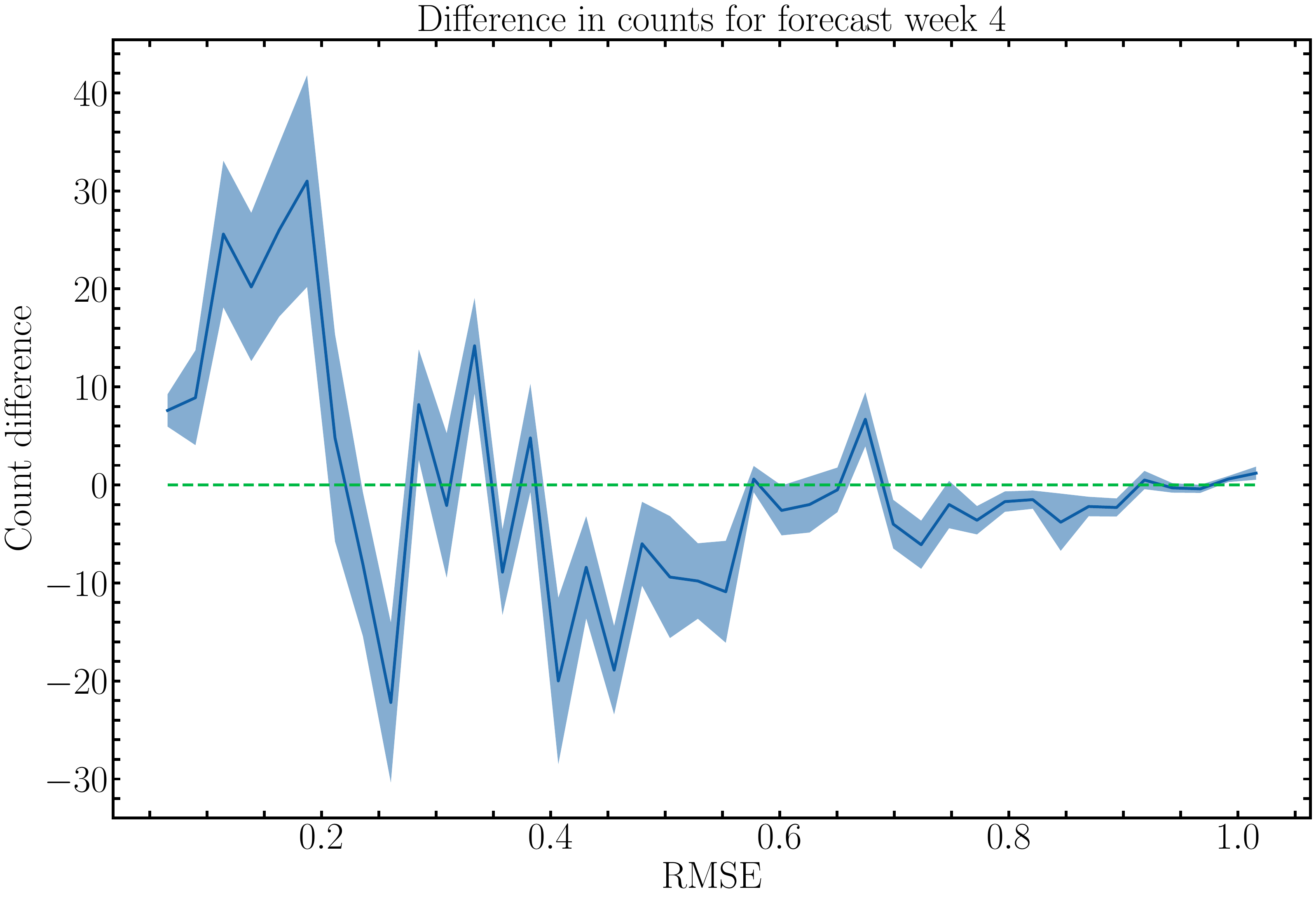}}
    } \\
    \mbox{
    \subfigure[Week 5]{\includegraphics[width=0.4\textwidth]{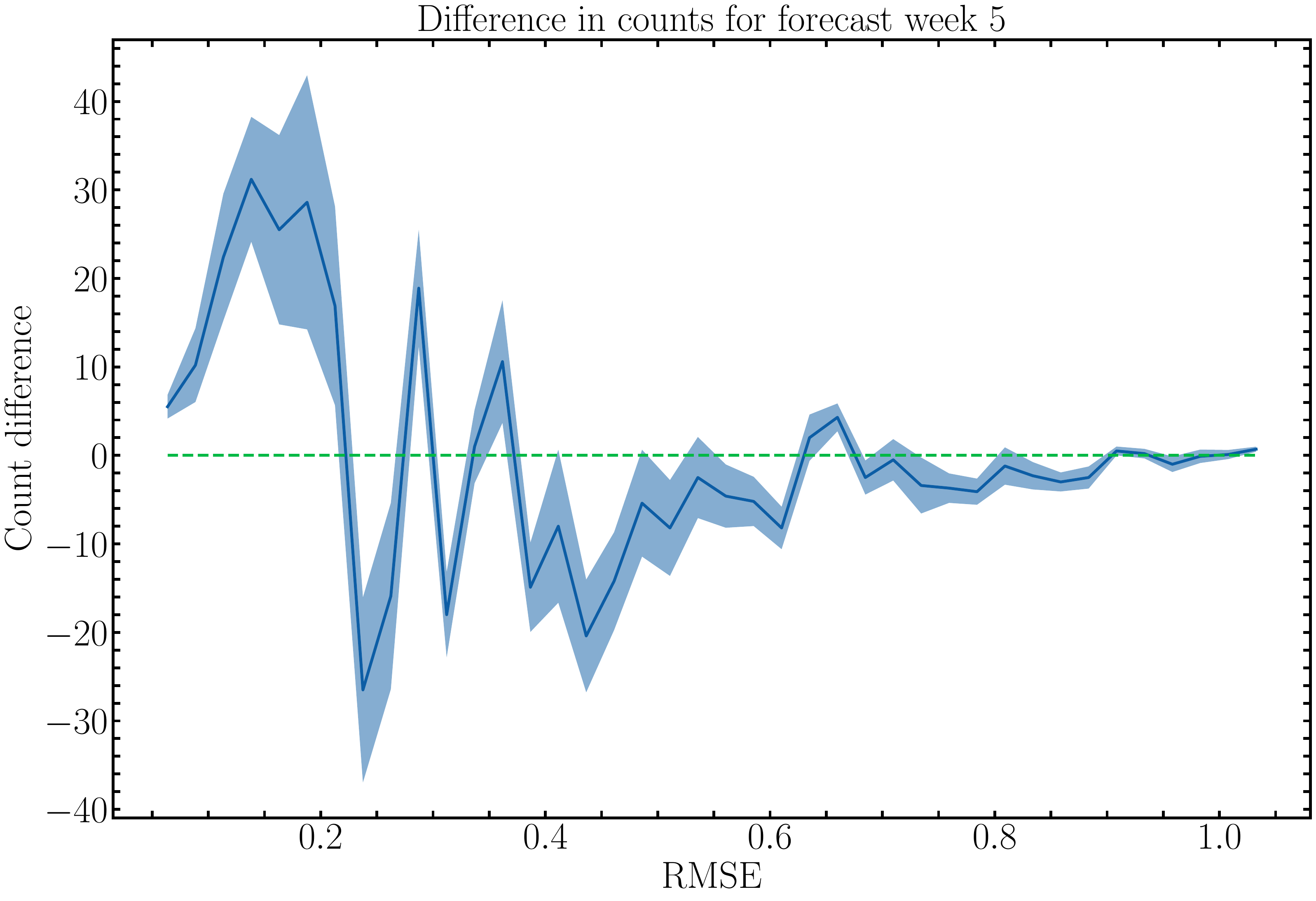}}
    \subfigure[Week 6]{\includegraphics[width=0.4\textwidth]{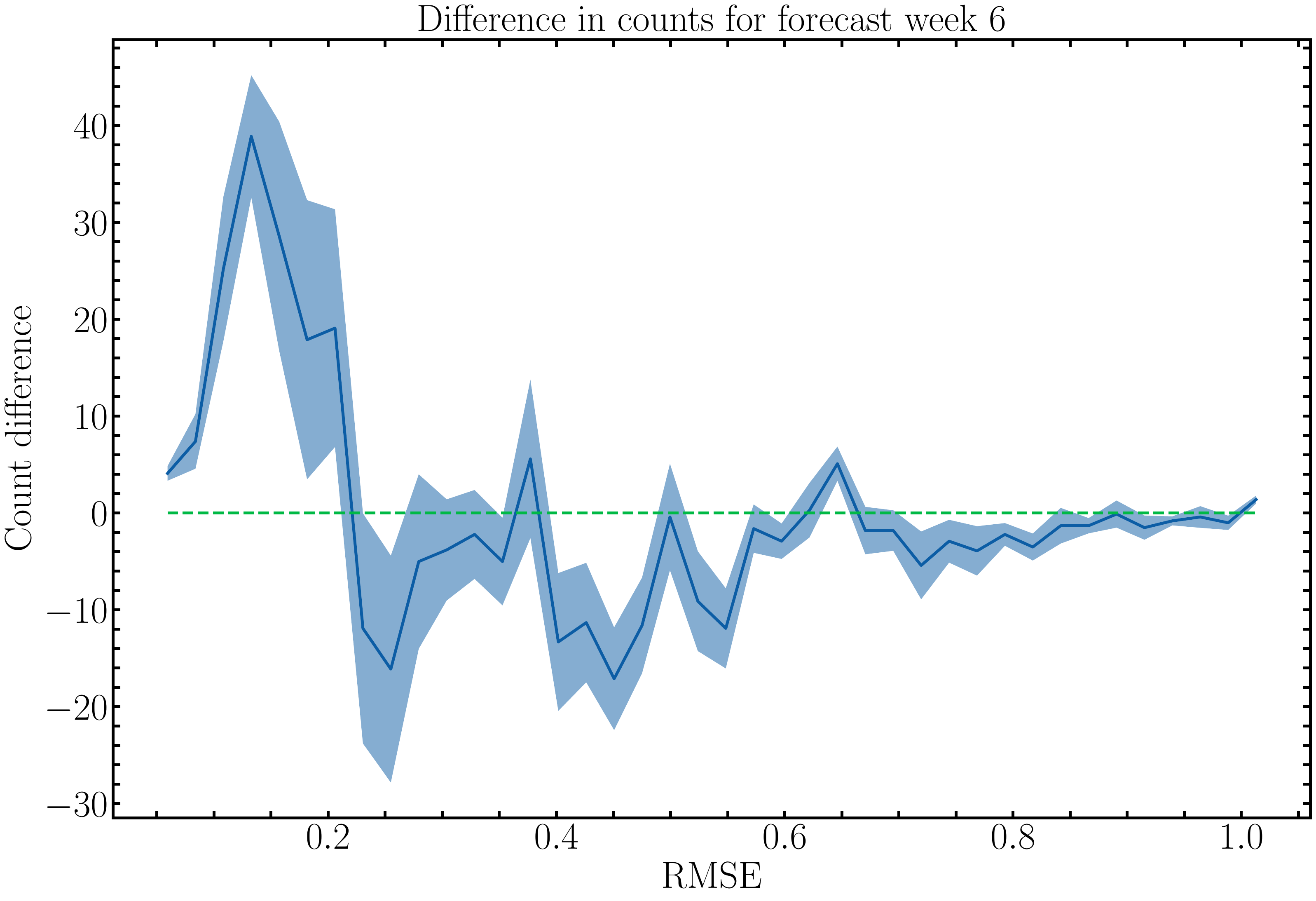}}
    } \\
    \mbox{
    \subfigure[Week 7]{\includegraphics[width=0.4\textwidth]{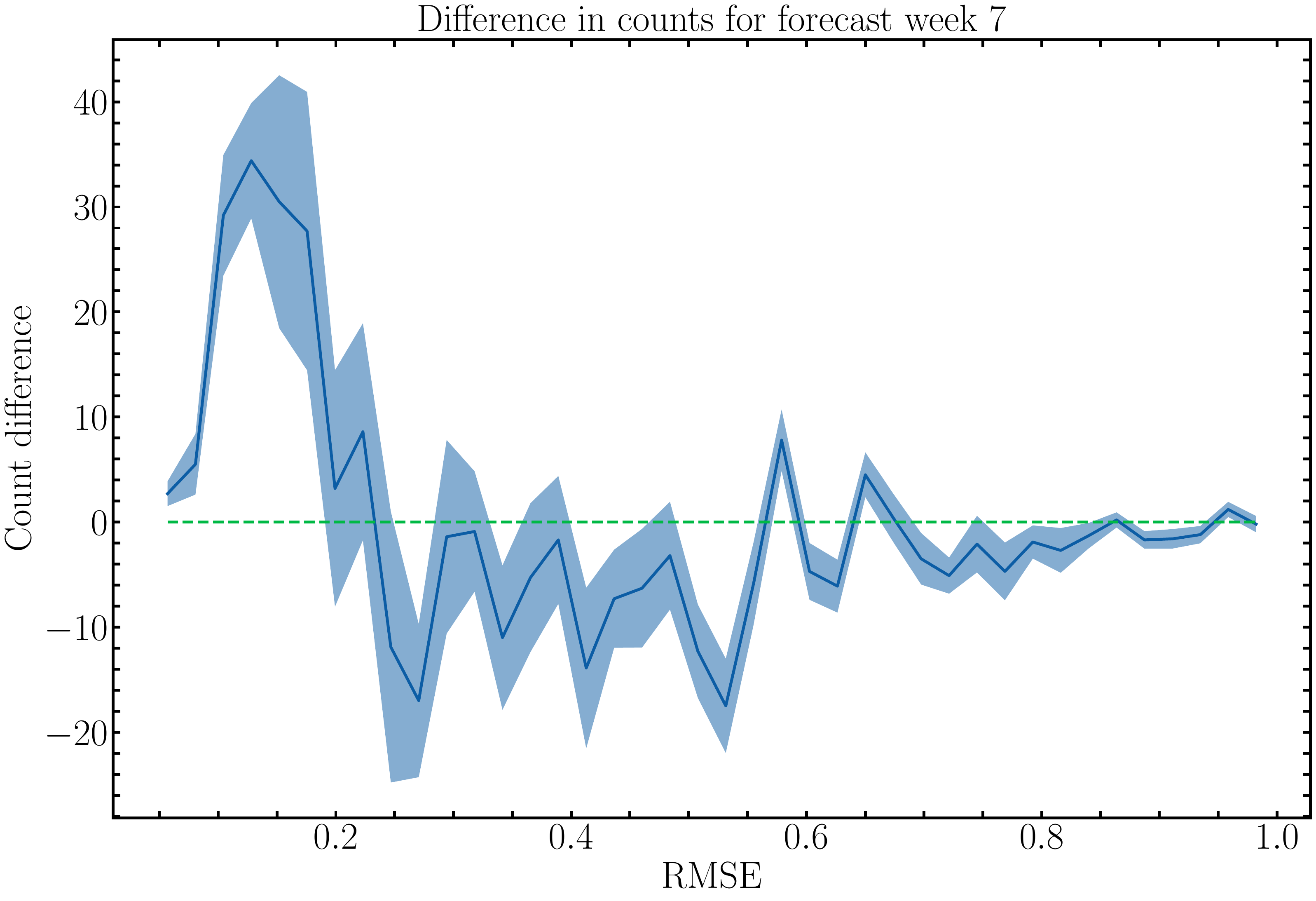}}
    \subfigure[Week 8]{\includegraphics[width=0.4\textwidth]{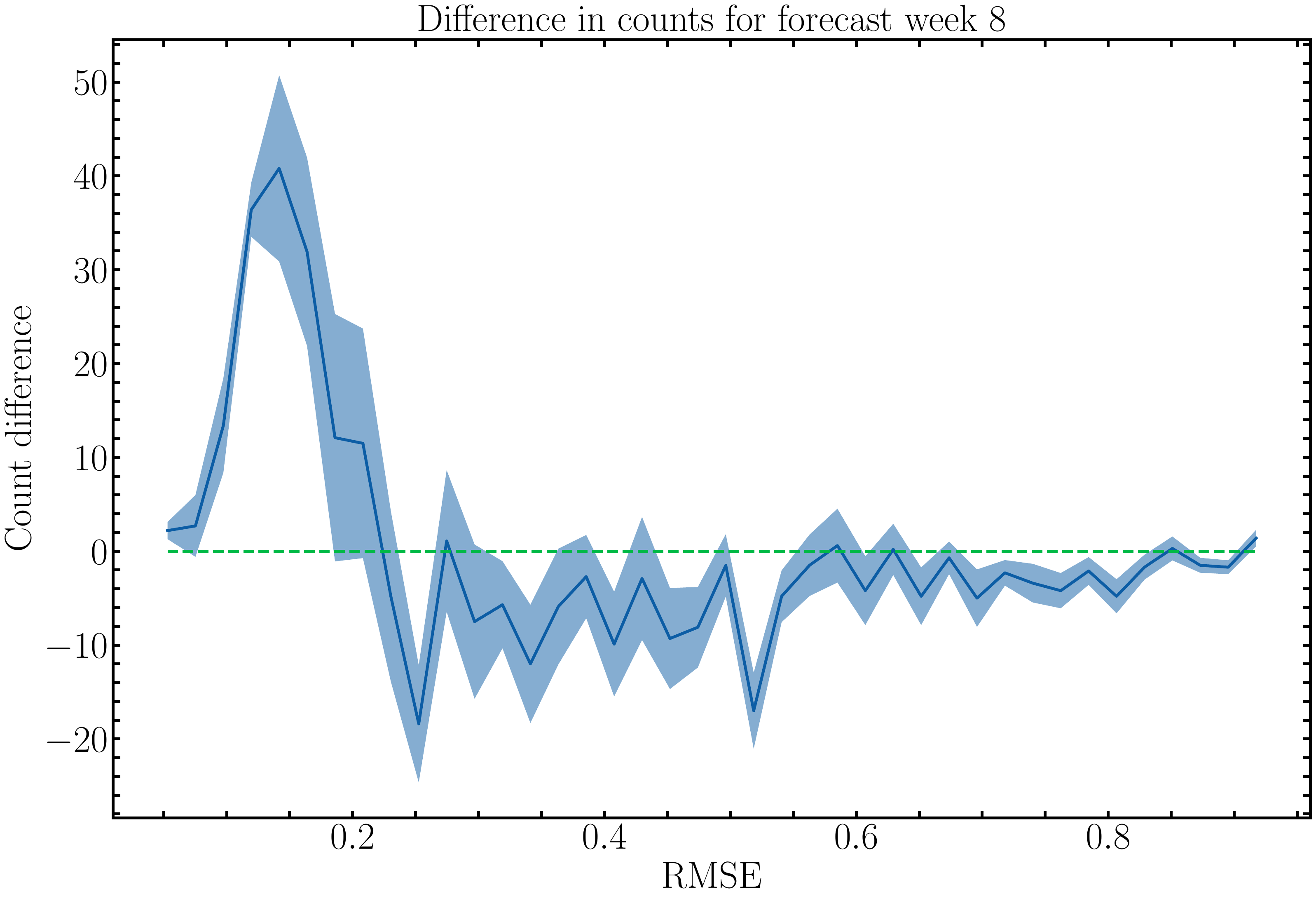}}
    }
    \caption{Showing count differences between the ensemble RMSE and individual member RMSE histograms for the SST forecasting task at different lead-times. Here count values larger than zero indicate a greater number of ensemble predictions at a particular RMSE level. It is clear that large positive values of this metric are encountered before large negative values for all weeks - indicating that the ensemble is superior to individual members.}
    \label{nas_ensemble_rmse_diff}
\end{figure}

\begin{figure}
    \centering
    \mbox{
    \subfigure[Epistemic Uncertainty]{\includegraphics[width=0.5\textwidth]{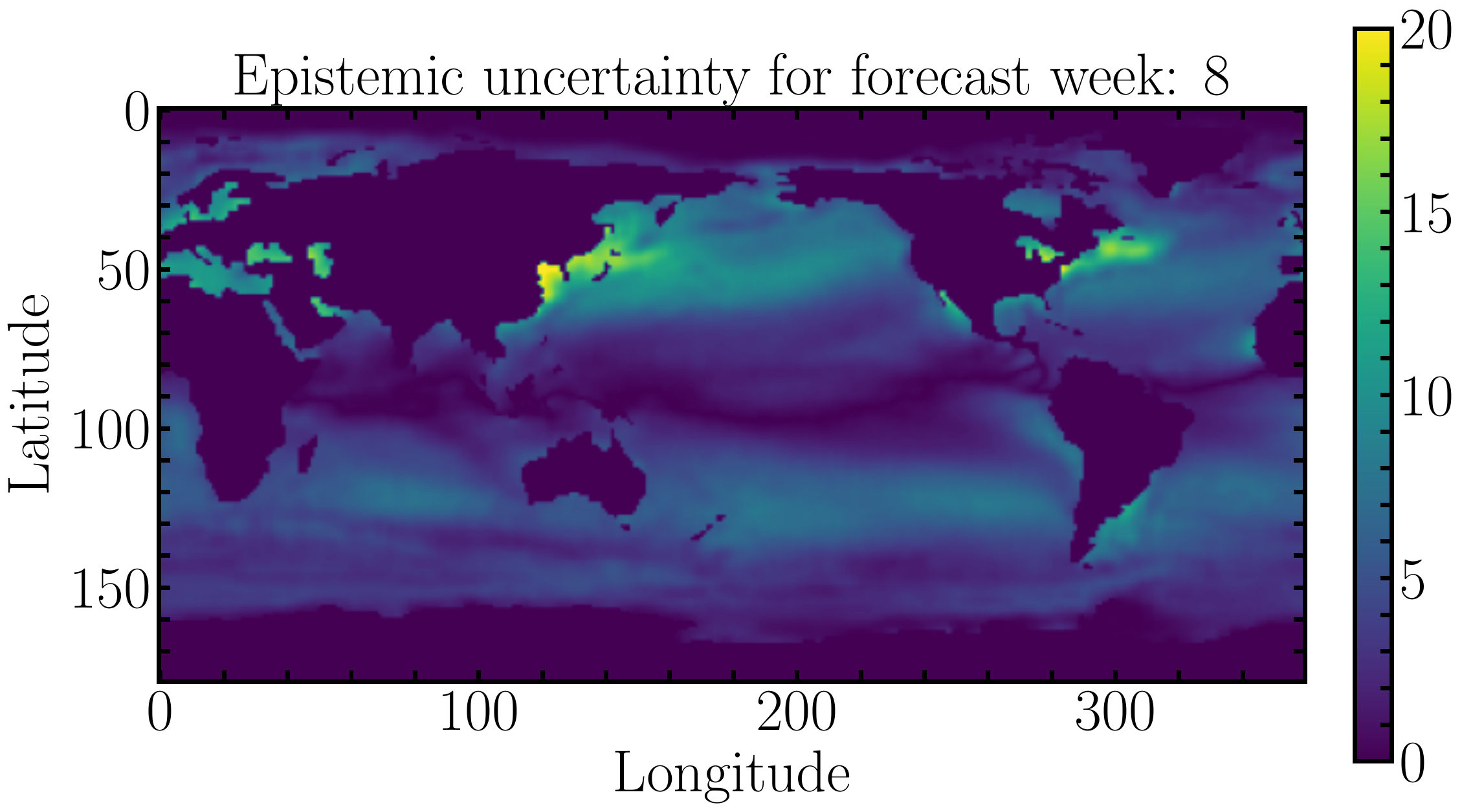}}
    \subfigure[Aleatoric Uncertainty]{\includegraphics[width=0.5\textwidth]{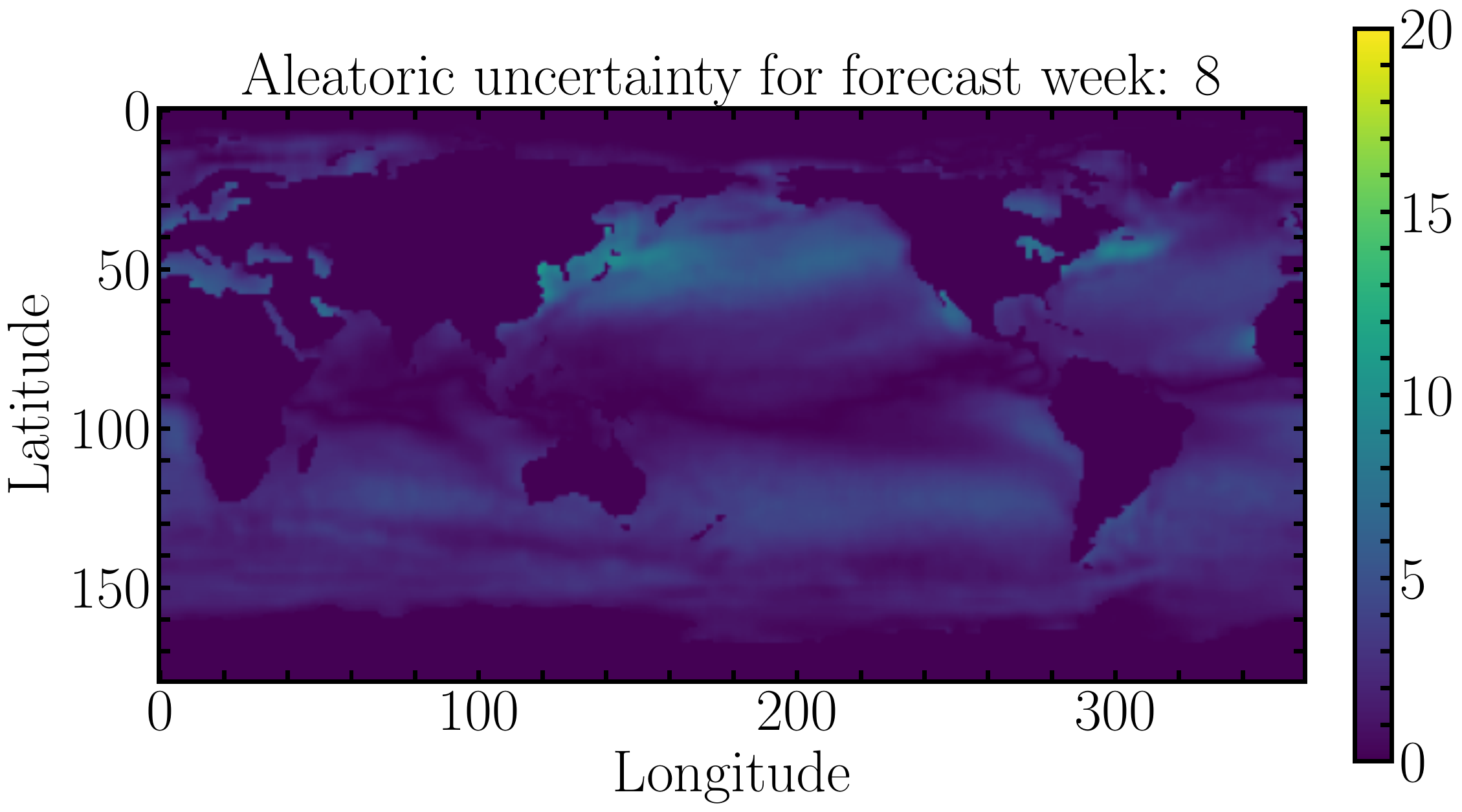}}
    }
    \mbox{
    \subfigure[Mean absolute error]{\includegraphics[width=0.5\textwidth]{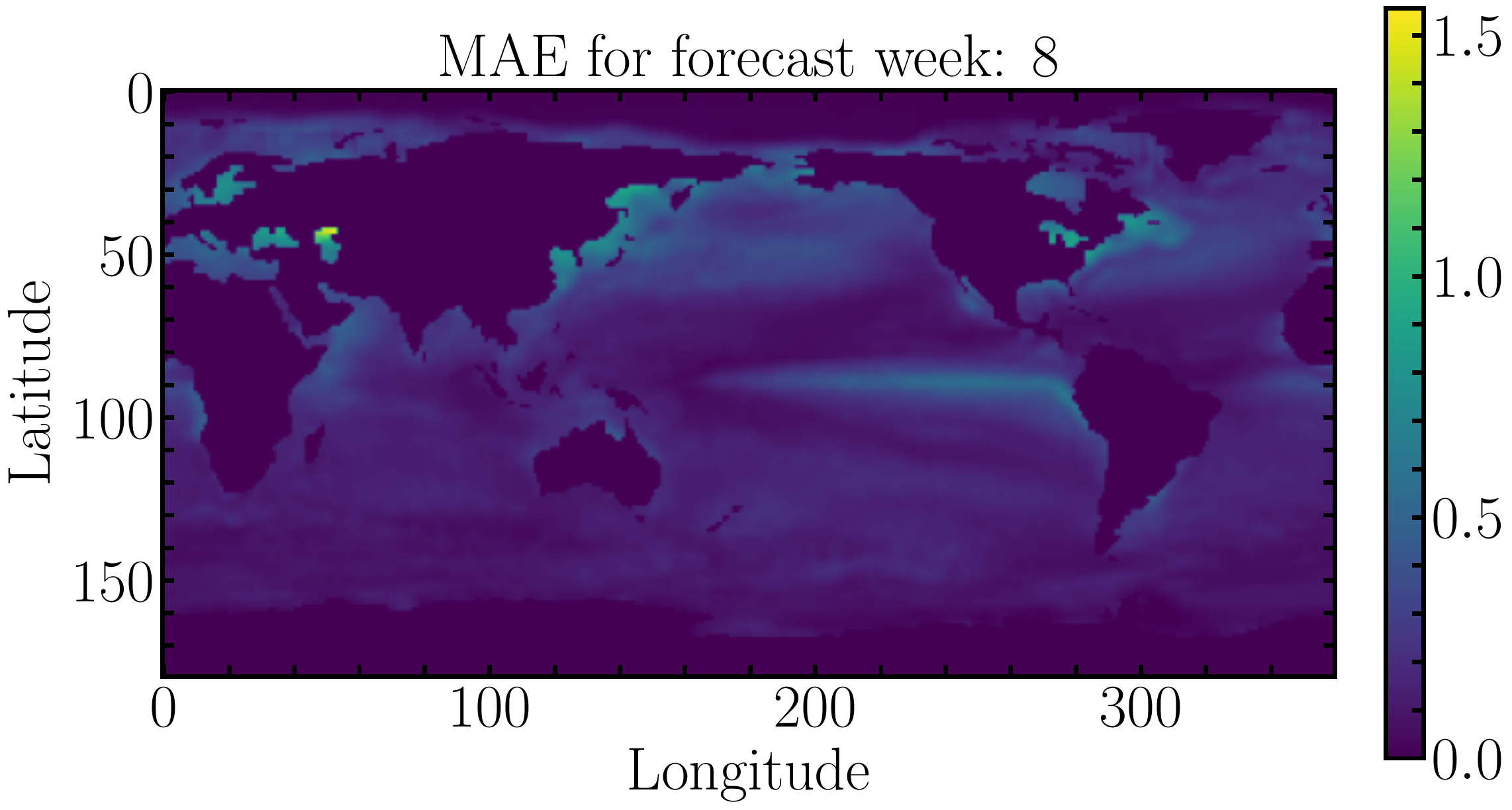}}
    
    }
    \caption{Epistemic uncertainty (top-left), aleatoric uncertainty (top-right) and mean absolute errors (bottom) for the forecasting problem (showing forecasts on week 8). Uncertainty is represented by the absolute value of the standard deviation of modal coefficients projected back into the physical space.}
    \label{nas_contours_lstm}
\end{figure}

Finally, we use the ensemble predictions from our 10 selected high-performing architectures to construct estimates of epistemic and aleatoric uncertainty from our identified surrogate models. Figure \ref{nas_contours_lstm} shows these plots for the final week of forecasts. A first takeaway from these plots is that Epistemic and Aleatoric uncertainty is characterized in coherent regions of the flow-field. These coherent regions are distinctly indicative of geographical trends. In terms of accuracy, the performance of the ensemble-based predictions (over here represented by the ensemble mean) improves on the previous search model reported in \cite{maulik2020recurrent} significantly as shown in Table \ref{RMSE_Table}. We also mention comparisons with two physics-based models given by the Community Earth System Model (CESM) \citep{kay2015community} \footnote{http://www.cesm.ucar.edu/projects/community-projects/LENS/data-sets.html} and Global Hybrid Coordinate Ocean Model (HYCOM) \footnote{https://www.ncdc.noaa.gov/data-access/model-data/model-datasets/navoceano-hycom-glb} that rely on the solution of complex partial differential equations on distributed computing resources. However, we highlight the fact that the slightly larger biases in the CESM and HYCOM data may be an artifact of interpolation from grids of different spatiotemporal resolution.

\begin{table*}[]
\centering
\caption{RMSE breakdown (in Celsius) for different forecast techniques compared against the NAS-POD-LSTM forecasts between April 5, 2015, and June 24, 2018, in the Eastern Pacific region (between -10 to +10 degrees latitude and 200 to 250 degrees longitude). Ensemble based forecasts outperform the `best-model' obtained via neural architecture search as reported in previous literature \citep{maulik2020latent}.}
\begin{tabular}{|c|c|c|c|c|c|c|c|c|}
\hline
\multicolumn{1}{|c|}{} & \multicolumn{8}{c|}{RMSE ($^\circ$Celsius) }\\
\hline
 & Week 1 & Week 2 & Week 3 & Week 4 & Week 5 & Week 6 & Week 7 & Week 8 \\ \hline
 This study & 0.33 & 0.34 & 0.34 & 0.35 & 0.37 & 0.38 & 0.38 & 0.39  \\ \hline
Maulik et al. \citep{maulik2020recurrent} & 0.62 & 0.63 & 0.64 & 0.66 & 0.63 & 0.66 & 0.69 & 0.65  \\ \hline
CESM      & 1.88 & 1.87 & 1.83 & 1.85 & 1.86 & 1.87 & 1.86 & 1.83  \\ \hline
HYCOM     & 0.99 & 0.99 & 1.03 & 1.04 & 1.02 & 1.05 & 1.03 & 1.05 \\ \hline
\end{tabular}
\label{RMSE_Table}
\end{table*}

From the perspective of computational cost, the search for the LSTM surrogate models in this section required 1 node-hour (8 GPU-hours) of wall-time on Theta GPU and evaluated 812 architectures sampled from the search-space. Here, it must be noted that for larger training requirements (i.e., when data sets are extremely large), smaller randomized subsets of the training data can be used to accelerate the convergence of architecture discovery before the full-data is exposed to the high-performing models. We leave such considerations for a future study. Additionally, our search procedure inherently priorities architectures that have fewer parameters since these members convergence more rapidly. In practice, this property of neural architecture search may be leveraged to find surrogates that lead to greater computational gains without sacrificing accuracy. Individual surrogate models identified are very cheap to deploy with each forward model evaluation obtained on the order of milliseconds.

\subsection{Reconstruction}

\begin{figure}
    \centering
    \includegraphics[width=0.5\textwidth]{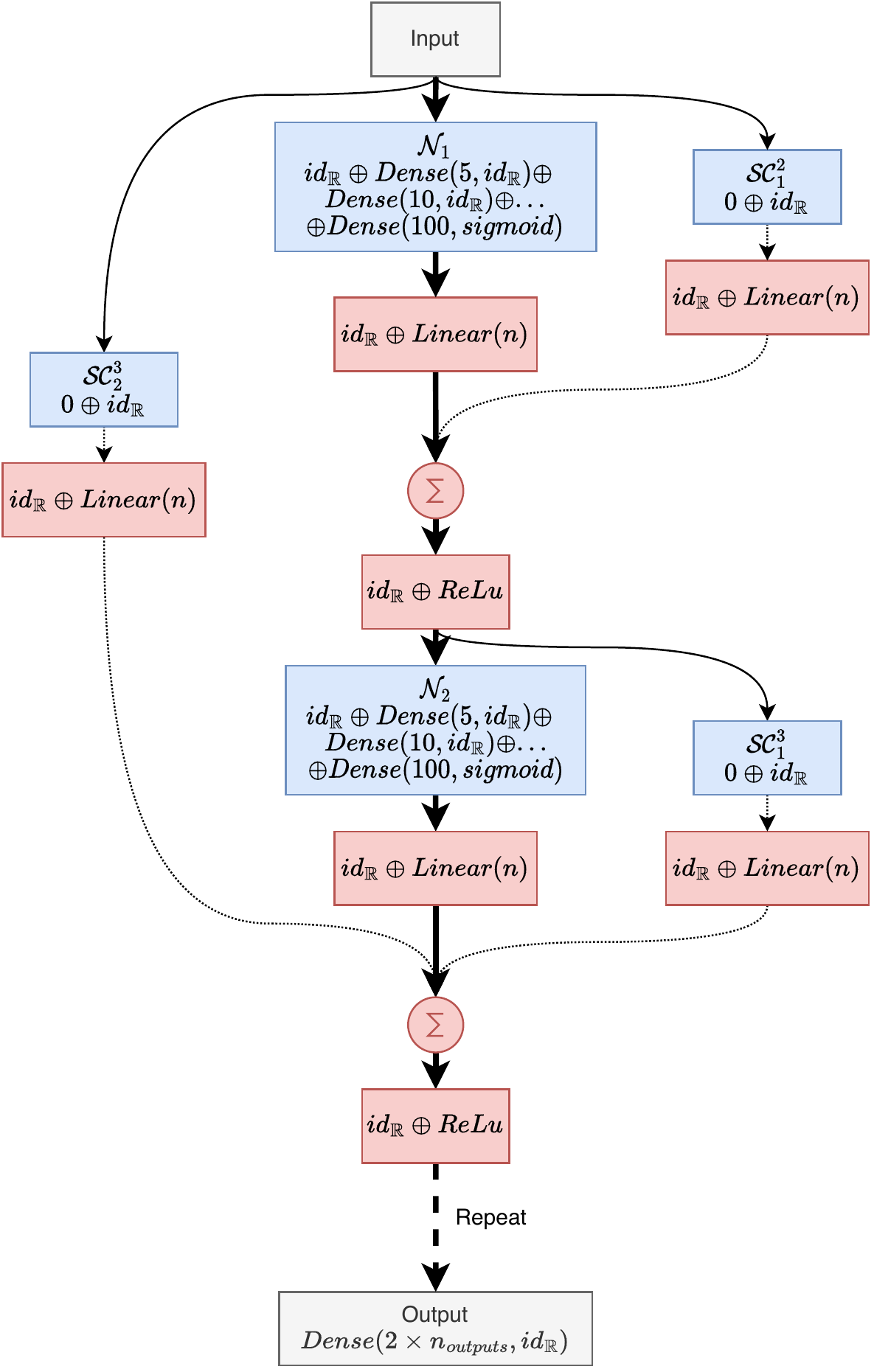}
    \caption{A description of the neural architecture search space constructed for the flow-reconstruction task.}
    \label{fig_space_rec}
\end{figure}

In this subsection, we evaluate our proposed methodology for the previously introduced benchmark flow-reconstruction problem. Our neural architecture search space builds a fully-connected network architecture from prespecified sensor input locations that measure the state, to the output which is the entire flow-field of dimension 360 times 180. The search space for sampling high-performing neural ensembles is shown in Figure. \ref{fig_space_rec}. In this search space, we have added 5 layers of fully-connected layers that are then allowed to interact within each other using discoverable skip connections. The training and testing data are segregated according to \cite{maulik2020probabilistic}, and specifically, the former is further split to obtain a validation data set that is used for directing the search for members of the ensemble.

\begin{figure}
    \centering
    \mbox{
    \subfigure[Convergence]{\includegraphics[width=0.5\textwidth]{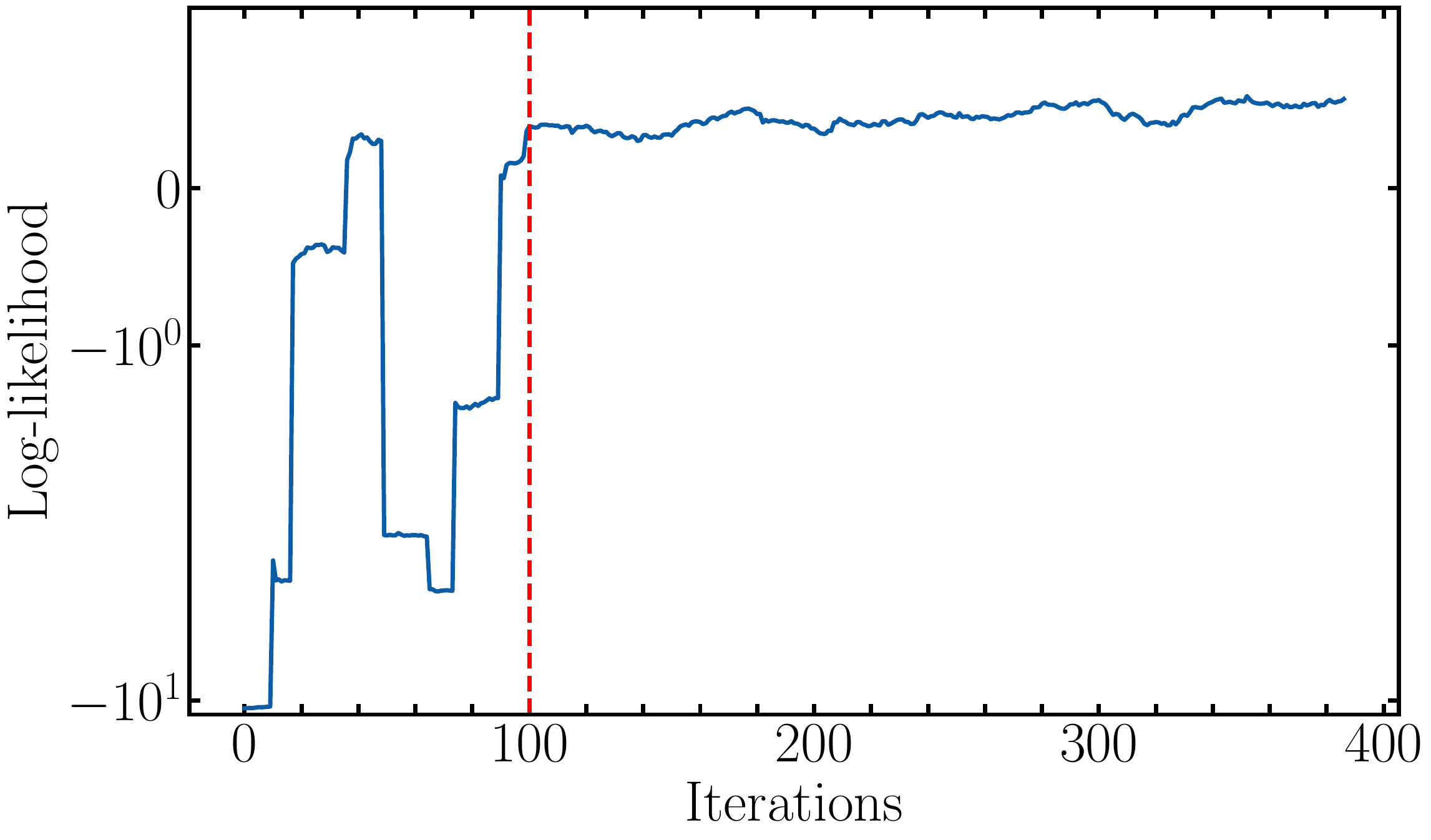} \label{nas_stats_fr_a}}
    \subfigure[Model spectrum]{\includegraphics[width=0.5\textwidth]{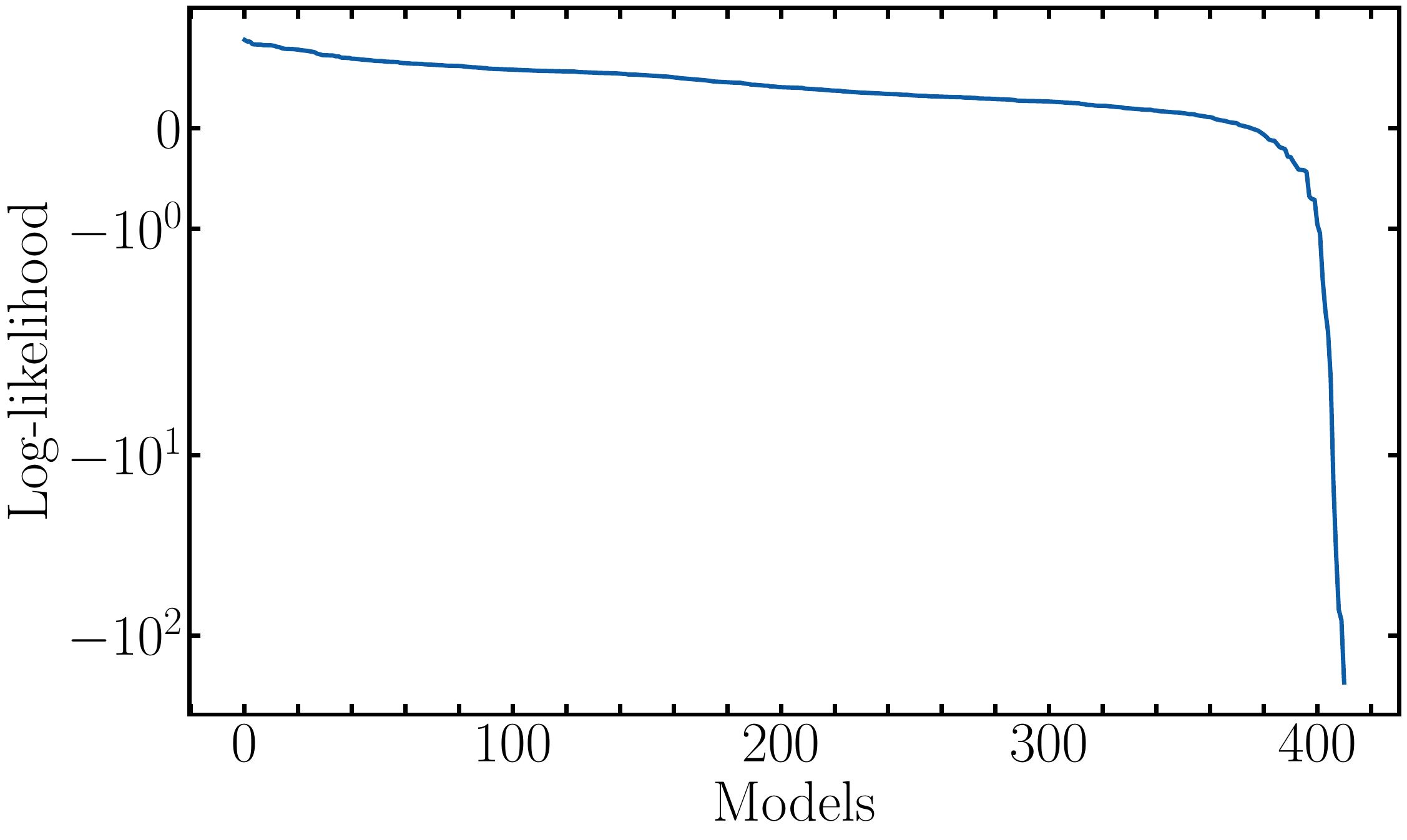}\label{nas_stats_fr_b}}
    }
    \caption{Convergence of neural architecture search (left) and spectrum of model performance (right) for the flow-reconstruction task. The vertical dashed line at 100 iterations indicates the start of intelligent exploration of the search space.}
\end{figure}

\begin{figure}
    \centering
    \mbox{
    \subfigure[RMSE histogram]{\label{nas_ensemble_fr}\includegraphics[width=0.49\textwidth]{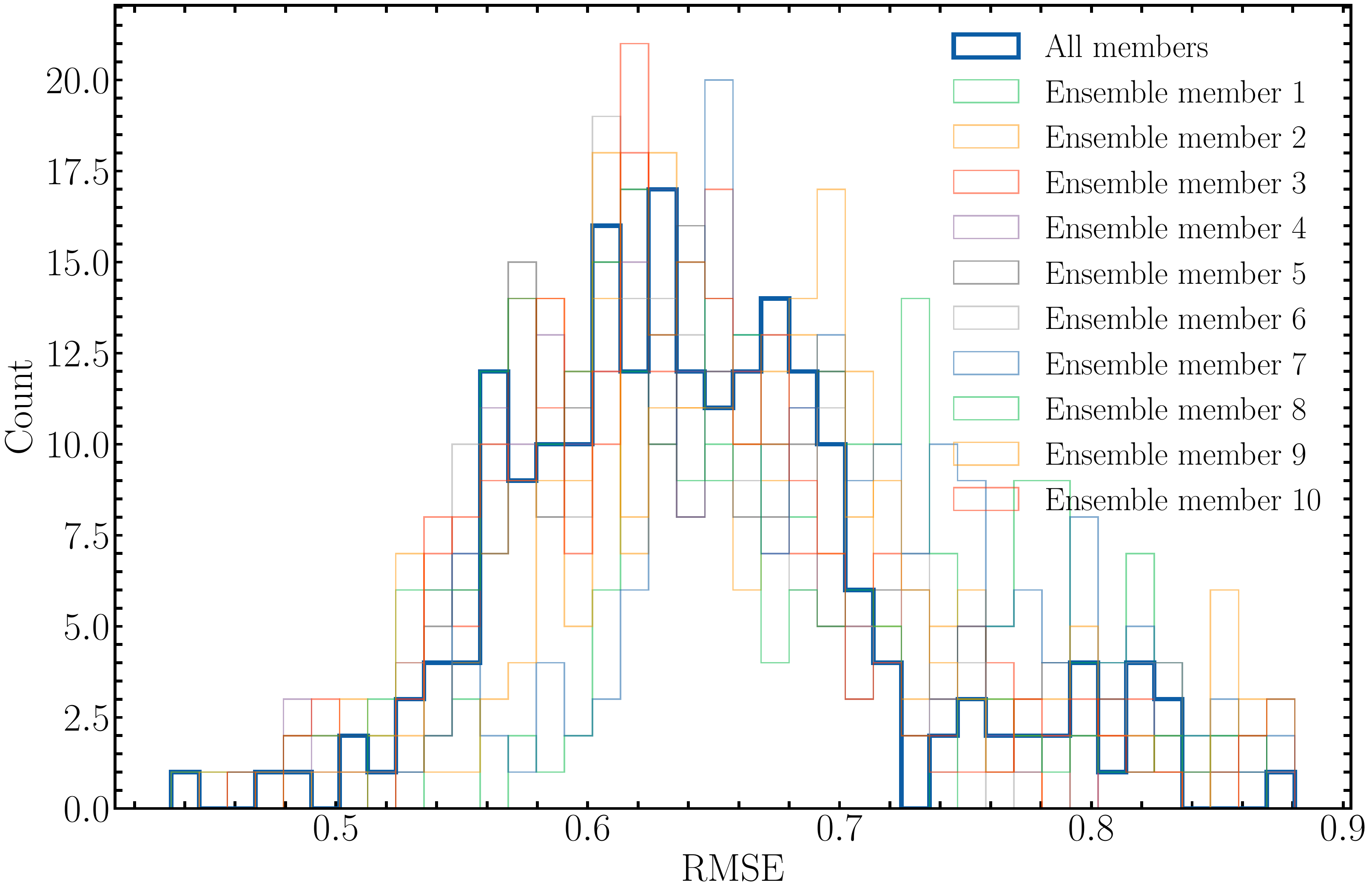}}
    \subfigure[RMSE count differences]{\label{nas_ensemble_fr_diff} \includegraphics[width=0.49\textwidth]{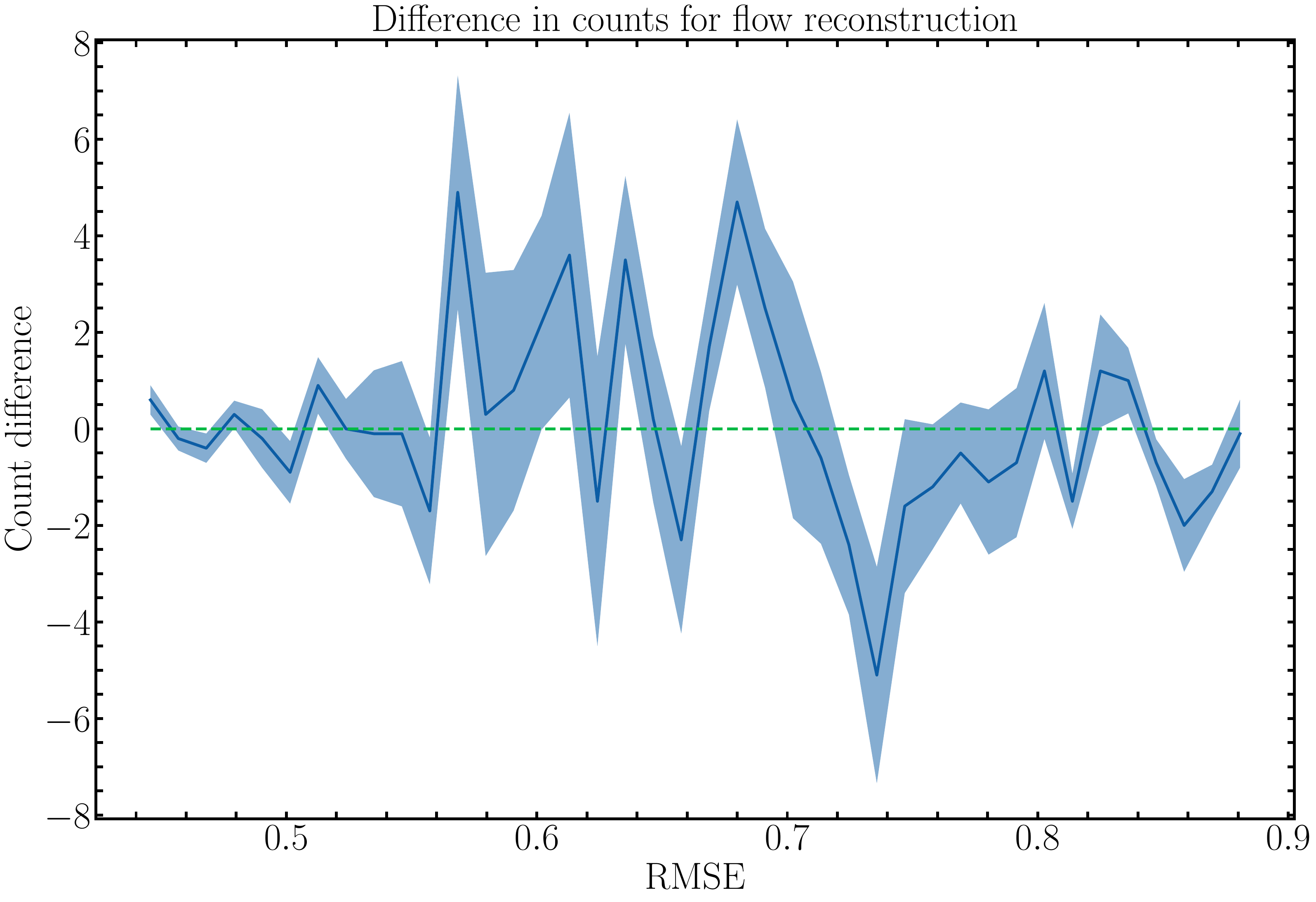}}
    }
    \caption{RMSE histogram for flow-reconstruction on test data set showing ensemble predictions as well as individual member statistics (left) and count differences between the ensemble RMSE and individual member RMSE for the SST flow reconstruction task. Here count values larger than zero indicate a greater number of ensemble predictions, i.e., there were more predictions from the ensemble with that value of an RMSE. It is clear that large positive values of this metric are encountered before large negative values - indicating that the ensemble is superior to individual members (right).}
\end{figure}

\begin{figure}
    \centering
    \mbox{
    \subfigure[Epistemic Uncertainty]{\includegraphics[width=0.5\textwidth]{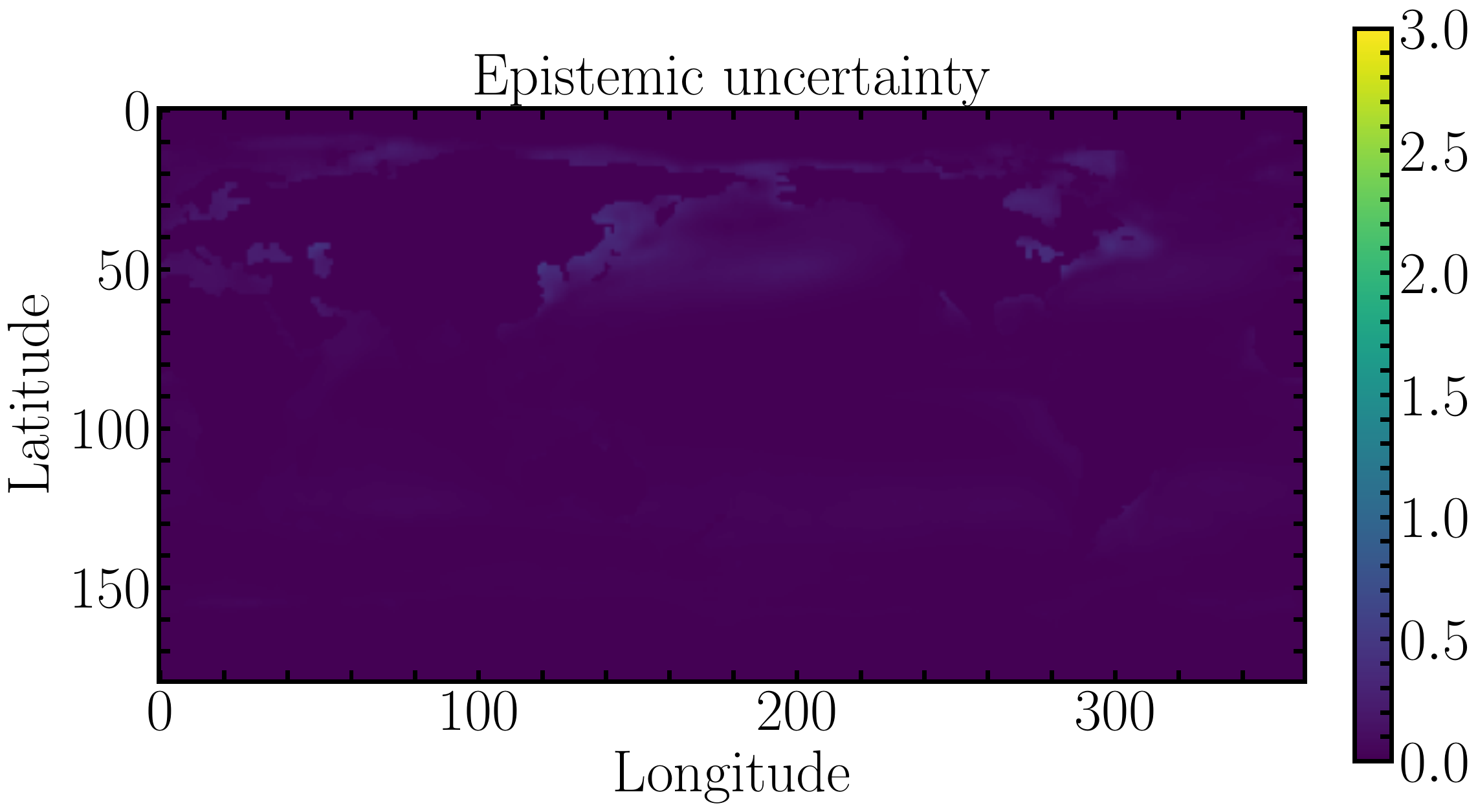}}
    \subfigure[Aleatoric Uncertainty]{\includegraphics[width=0.5\textwidth]{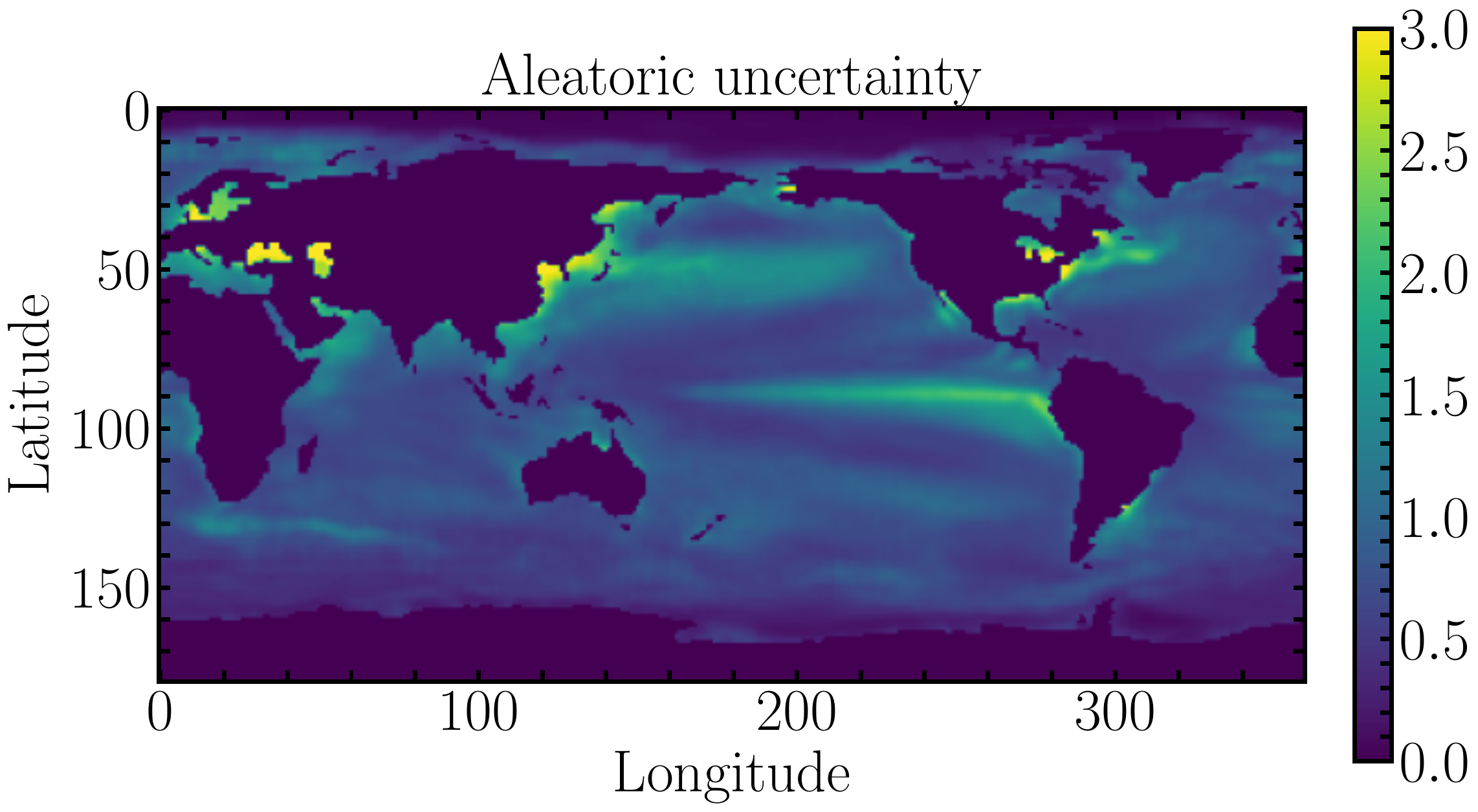}}
    }
    \mbox{
    \subfigure[Mean absolute error]{\includegraphics[width=0.5\textwidth]{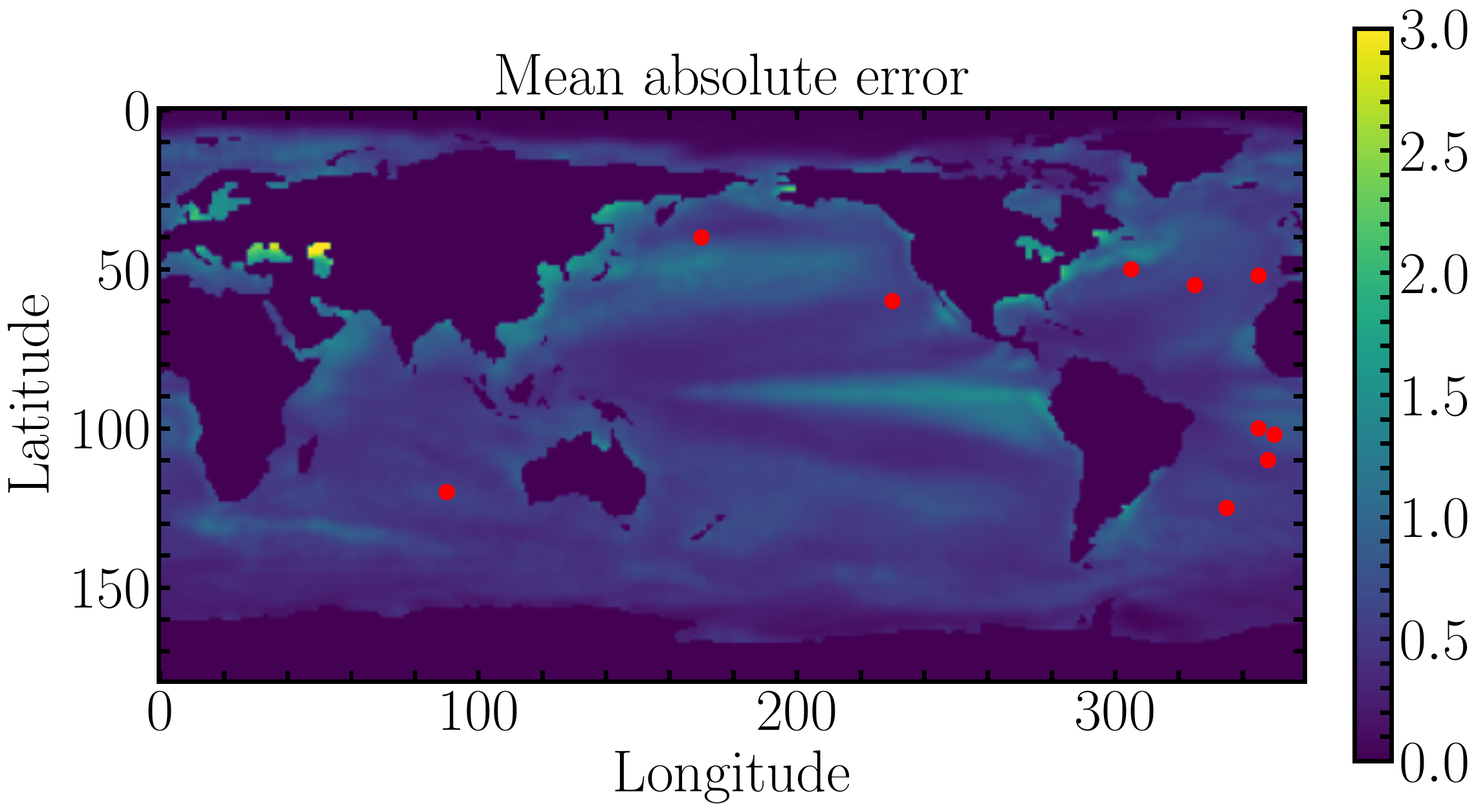}}
    }
    \caption{Epistemic uncertainty (top-left), aleatoric uncertainty (top-right) and mean absolute errors (bottom) for the flow-reconstruction problem. The sensor locations are shown in red. One can observe that the mean-absolute errors are highly correlated with the aleatoric uncertainty.}
    \label{nas_contours_fr}
\end{figure}

Figure \ref{nas_stats_fr_a} shows the convergence of the neural architecture search for the flow-reconstruction task. We observe an initial period of exploration which leads to several underperforming architectures but the search is soon seen to converge to a high value of the log-likelihood. Figure \ref{nas_stats_fr_b} shows the range of model performance for all the sampled architectures during the search. A small proportion of the total models is seen to have poor performance, representing the initial stage of exploration by the differential evolution algorithm. As in the previous example, we retain the 10 best models obtained at the end of this search and compute statistics for RMSE for individual ensemble members as well as the prediction by the ensemble. This is shown through histogram counts in Figure \ref{nas_ensemble_fr}, where one can qualitatively observe that large counts in the tails in some members of the ensemble are reduced. The improvement on individual members of the ensemble can be observed in Figure \ref{nas_ensemble_fr_diff} where larger values of the difference in counts imply that the ensemble made more predictions with that value of the error. The desirable behavior of have large positive values at low RMSE and larger negative values at high RMSE is observed here as well. Finally averaged estimates of uncertainty are shown in Figure \ref{nas_contours_fr} where it can be seen that epistemic uncertainty is mainly driven by coastal areas in the northern hemisphere whereas aleatoric uncertainty arises from the typical ENSO fluctuations in the Pacific Ocean. The metric used for assessing the performance of the ensemble is given by the relative $L_2$-norm error of 0.0428 which is comparable to the model obtained in \cite{maulik2020probabilistic} but with added estimates of epistemic and aleatoric uncertainty represented clearly. From a computational perspective, the neural architecture search was performed using 8 GPU hours, with the total evaluation of 401 models and the model evaluation post-training was performed virtually instantaneously. A significant correlation between the error and the uncertainty is also clearly observed.

\section{Conclusion}

In this paper, we have demonstrated a computational workflow that may be used both for constructing deep learning models for forecasting dynamical systems and for reconstructing flow-field information from sparse sensors.  Our experiments demonstrate that a joint neural architecture and hyperparameter search not only discovers effective models for target function approximation but also doubles as an effective tool to quantify uncertainty in our predictions. This is accomplished by using a number of the high-performing models discovered by the deep neural network search in an ensemble-based uncertainty quantification approach. By endowing our search space with an output layer that may use a maximum-likelihood based estimation, i.e., if the outputs are both the mean as well as variance which goes into a Gaussian likelihood maximization formulation, we are able to quantify the aleatoric uncertainty of the predictions associated with each model. By leveraging the law of decomposition of the total uncertainty of an ensemble, we are also able to quantify the aleatoric and epistemic uncertainty in predictions of the entire ensemble of members. Our data set for experiments comes from the NOAA optimum interpolation sea-surface temperature data set, thereby representing a representative data set for complex geoscience data set. Our conclusions suggest that scalable distributed neural architecture and hyperparameter search may represent a vertically integrated workflow for high-performing surrogate model construction as well as efficient uncertainty quantification.

\section*{Data availability}

Training data used in the present study are available publicly on Google Drive: \\
Example 1 (NOAA sea surface temperature - forecasting): \\ \url{https://drive.google.com/drive/folders/1pVW4epkeHkT2WHZB7Dym5IURcfOP4cXu?usp=sharing}), \\
Example 2 (NOAA sea surface temperature - reconstruction): \\ \url{https://drive.google.com/drive/folders/1pVW4epkeHkT2WHZB7Dym5IURcfOP4cXu?usp=sharing}),

\section*{Code availability}

Sample codes for neural architecture search as well as ensemble based predictions will be made available after peer-review (or on request). Tutorials and examples for deploying the neural architecture and hyperparameter searches for arbitrary modeling tasks can be found at \texttt{https://github.com/deephyper/deephyper}.

\section*{Acknowledgement}

This work was supported by the U.S. Department of Energy, Office of Science, Office of Advanced Scientific Computing Research, under Contract~DE-AC02-06CH11357. We acknowledge funding support from ASCR for DOE-FOA-2493 ``Data-intensive scientific machine learning" and the DOE Early Career Research Program award. This research was funded in part and used resources of the Argonne Leadership Computing Facility, which is a DOE Office of Science User Facility supported under Contract DE-AC02-06CH11357.



\section{Declaration of interest}

The authors report no conflict of interest.

\bibliographystyle{unsrtnat}
\bibliography{references}







\end{document}